\documentclass[11pt]{article}
\usepackage[preprint]{acl}
\usepackage{times}
\usepackage{latexsym}
\usepackage[T1]{fontenc}
\usepackage[utf8]{inputenc}
\usepackage{microtype}
\usepackage{inconsolata}
\usepackage{amsmath,amssymb,booktabs,tabularx,array}
\usepackage{graphicx}
\usepackage{colortbl}
\definecolor{tableblue}{HTML}{EFF3F6}
\definecolor{tablepeach}{HTML}{E7EEF5}

\usepackage{longtable}
\usepackage{algorithm,algpseudocode}
\usepackage{fvextra}
\usepackage{placeins}

\usepackage[skins]{tcolorbox}

\definecolor{choiceink}{HTML}{26324D}
\definecolor{choicepaper}{HTML}{F7F7F9}

\tcbset{choicecallout/.style={
  enhanced,
  colback=choicepaper,
  colframe=choiceink,
  boxrule=0.45pt,
  arc=1.2pt,
  left=4pt,
  right=4pt,
  top=6.5pt,
  bottom=2pt,
  before skip=8pt,
  after skip=6pt,
  fontupper=\normalsize,
  fonttitle=\normalsize\bfseries,
  coltitle=white,
  attach boxed title to top left={
    xshift=5pt,
    yshift=-7pt
  },
  boxed title style={
    colback=choiceink,
    colframe=choiceink,
    boxrule=0pt,
    arc=1pt,
    left=4pt,
    right=4pt,
    top=2pt,
    bottom=2pt
  }
}}

\newtcolorbox{researchquestionbox}{
  choicecallout,
  title=Research Question
}
\newtcolorbox{keyfindingsbox}{
  choicecallout,
  title=Key Findings
}
\newtcolorbox{takeawaybox}{
  choicecallout,
  title=Takeaway
}
\definecolor{findingink}{HTML}{26324D}
\newtcolorbox{findingbox}[1]{choicecallout,title={#1}}
\definecolor{promptslot}{HTML}{17619A}
\newtcolorbox{promptbox}[1]{enhanced,
  colback=white,colframe=findingink,colbacktitle=findingink,
  coltitle=white,fonttitle=\bfseries\small,boxrule=0.45pt,arc=0.7mm,
  left=2mm,right=2mm,top=2mm,bottom=2mm,title={#1},
  before skip=6pt,after skip=6pt}
\newcommand{\promptfield}[1]{\textcolor{promptslot}{\{#1\}}}
\newcommand{\lead}[1]{\noindent\textbf{#1}\ }
\newcommand{\quadacc}{\mathrm{QAcc}}

\newcommand{\ind}{\mathbf{1}}
\newcolumntype{L}{>{\raggedright\arraybackslash}X}

\title{Which Differences Matter?\\
\large\itshape Controlled Tests of Rule Transfer, Joint Reliability,\\
and Information Use in Decision Models}
\author{
  Tianyou Lai\textsuperscript{1}\thanks{These authors contributed equally to this work.}\quad
  Wentao Yue\textsuperscript{2}\footnotemark[1]\quad
  Zhiwei Zuo\textsuperscript{3}\quad
  Qilei Li\textsuperscript{3}\thanks{Corresponding author: \href{mailto:qilei.li@ccnu.edu.cn}{\texttt{qilei.li@ccnu.edu.cn}}.}\\
  \normalfont\small\textsuperscript{1}Carnegie Mellon University\quad
  \textsuperscript{2}Shenzhen University\\
  \normalfont\small\textsuperscript{3}Central China Normal University
}
\hypersetup{
  pdftitle={Which Differences Matter? Controlled Tests of Rule Transfer, Joint Reliability, and Information Use in Decision Models},
  pdfauthor={Tianyou Lai, Wentao Yue, Zhiwei Zuo, Qilei Li}
}

\begin{document}
\raggedbottom
\maketitle

\begin{abstract}
Decision models map stated rules and observed states to discrete answers. Aggregate accuracy leaves unresolved which rule and information changes guide those answers. We connect controlled input changes to transfer, joint reliability, and directed information use through executable arithmetic rules. Semantic rule separation and crossed rule and state tests make transfer and joint success separately measurable. We compare equivalent wordings, calibrate joint correctness through marginal-preserving reassociation, and alter tables to define alternative answer targets. Adaptation gains increase with nearest-source answer agreement; fully conflicting quartets yield small pooled gains and different gain signs across source splits. Equivalent rewrites have opposing task-level effects, linking transfer gains to rule presentation. Joint-score departures depend on which rule and state associations the reference preserves. A decision service changes answers more across equivalent wordings than across identical-text repeats and usually follows the answer targets implied by altered table values or mappings. Code and data will be made available.
\end{abstract}

\begin{figure*}[!t]
\centering
\includegraphics[width=\textwidth]{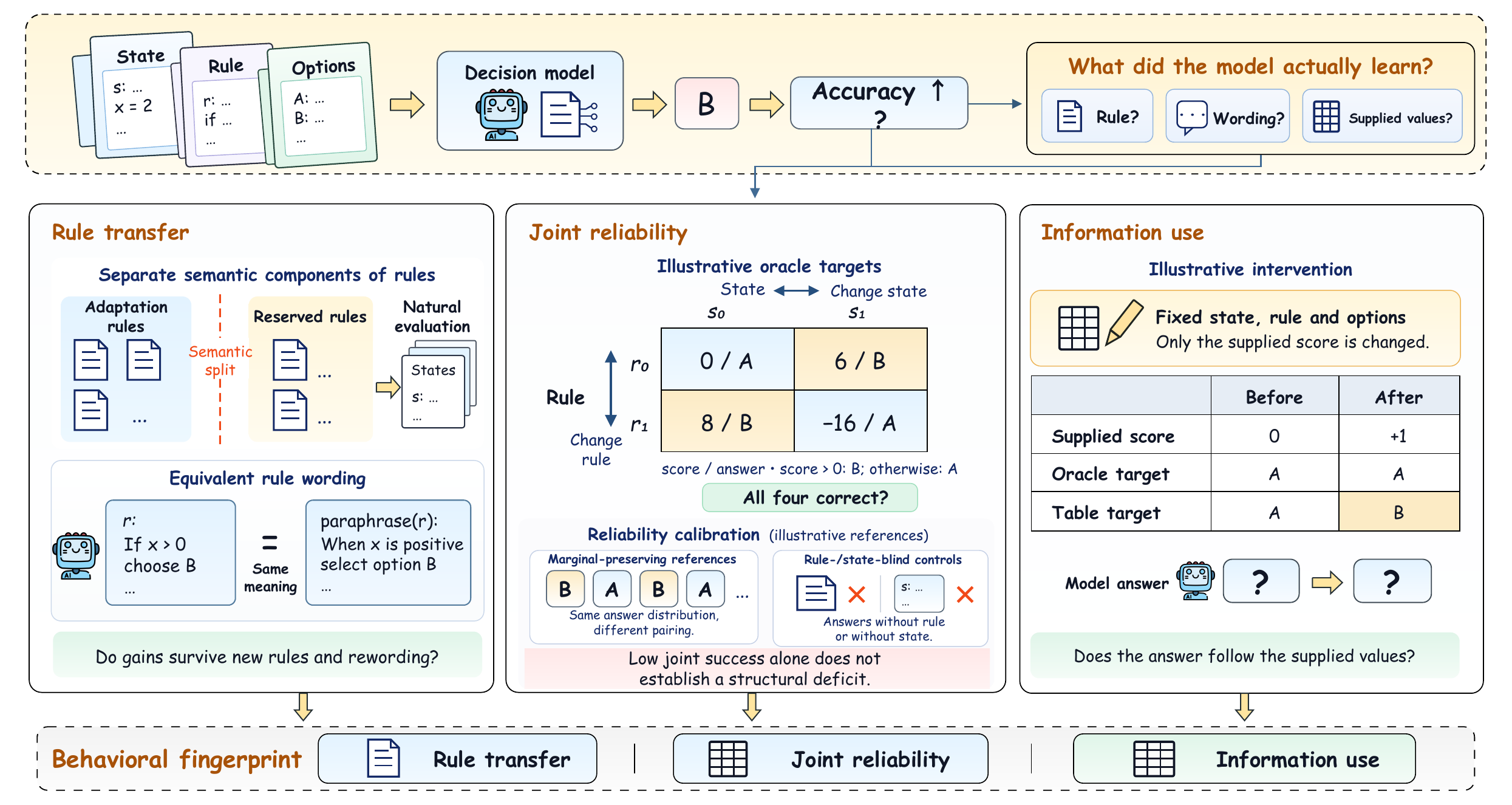}
\caption{Overview of the three behavioral diagnostics: rule transfer and wording, joint reliability, and external information use. Scores and interventions illustrate the design.}
\label{fig:study-overview}
\end{figure*}

\section{Introduction}
\begin{flushright}
\small
\textit{``A difference which makes a difference''}\\[2pt]
{\footnotesize\mbox{\textemdash{} Gregory Bateson, \textit{Form, Substance and Difference} (1970)}}
\end{flushright}
Decision models turn stated rules and observed states into labels, ratings, or selected records. Accuracy measures agreement with the prescribed answer, but equal scores can conceal different responses to the inputs. In an archived routing example, two models answer half the questions correctly. One ignores both rule and state changes; the other responds to states alone. Valid option selection can also coexist with answer-symbol binding errors \citep{balepur-etal-2025-best,xue-etal-2024-strengthened}. Understanding success therefore requires testing which input changes guide each rule-based decision.

\begin{researchquestionbox}
How can we determine which rule and information changes guide a decision model\textquotesingle{}s answers?
\end{researchquestionbox}

A gain on unfamiliar rules can still depend on familiar wording. Low all-four correctness can follow from individual errors, so its magnitude alone does not establish unusual joint behavior. Better performance with an external table also leaves open whether its values and mappings direct the answers. Behavioral tests and contrast sets motivate controlled input changes \citep{ribeiro-etal-2020-beyond,gardner-etal-2020-evaluating}; rule-following benchmarks examine inference, counterfactual rules, and regulations \citep{parmar-etal-2024-logicbench,sun-etal-2025-beyond-instruction,zhou-etal-2025-rulearena}. We connect these controls to the specific ambiguities in interpreting decision accuracy.

Executable arithmetic rules make intended answers and controlled changes checkable. We separate rules by finite-domain behavior to evaluate transfer to semantic components excluded from training. Equivalent rewrites hold those behaviors fixed while changing presentation. Crossing rules and states tests responses to each factor. Marginal-preserving reassociation calibrates all-four correctness against the observed individual predictions. More restrictive references retain rule or state associations to identify what drives departures from that calibration. Finally, changing table values or mappings defines alternative answer targets. Transitions toward those targets reveal directed information use beyond an arbitrary answer change. Figure~\ref{fig:study-overview} links input changes to behavioral diagnostics.

Low-rank adaptation (LoRA) improves accuracy on reserved rules, with larger gains at higher nearest-source answer agreement. Fully conflicting quartets have low joint success and opposite gain signs across source splits. Equivalent wording increases the adaptation advantage on binary decisions and reduces it on the other tasks. Joint-score departures further depend on the rule and state associations retained by the calibration reference. The decision service Jev changes answers more across equivalent wordings than across identical-text repeats. It benefits from complete value tables and follows the alternative targets implied by altered values and mappings. These diagnostics distinguish conflict-conditioned transfer, joint reliability, and directed information use in model decisions.

\begin{itemize}
\item We establish an oracle-checked evaluation that uses semantic rule separation and paired input changes to distinguish transfer, wording sensitivity, and responses to rules and states.
\item We show that transfer gains depend on source-answer agreement and the source split under complete conflict. Joint-score departures depend on the associations retained by different marginal-preserving calibration references.
\item We identify directed use of external values through paired answer transitions under value and mapping interventions, complementing the evidence from aggregate accuracy gains.
\end{itemize}

\section{Related Work}
\paragraph{Behavioral evaluation and explicit rules.}
Behavioral evaluation tests whether predictions respond correctly to controlled input changes. CheckList and contrast sets target capabilities through local perturbations \citep{ribeiro-etal-2020-beyond,gardner-etal-2020-evaluating}. Metric rankings also depend on the evaluation context \citep{deviyani-diaz-2025-contextual}. RuleTaker and ProofWriter evaluate inference from stated facts and rules \citep{clark2020ruletaker,tafjord-etal-2021-proofwriter}. LogicBench spans logical systems; RuleArena separates rule selection from application in regulations \citep{parmar-etal-2024-logicbench,zhou-etal-2025-rulearena}. RuleBench and IRFT extend inferential rule-following evaluation and training \citep{sun-etal-2025-beyond-instruction}. These designs motivate testing which inputs guide decisions. Our executable grammar combines finite-domain semantic separation with crossed rule and state changes, linking success to responses along each input axis. Appendix Table~\ref{tab:literature-comparison} compares these paired evaluation designs.

\paragraph{Generalization, wording, and answer interfaces.}
Generalization benchmarks vary combinations, presentations, and answer interfaces. CFQ and COGS test composition; Ordered CommonGen tests requested concept order \citep{keysers2020measuring,kim-linzen-2020-cogs,sakai-etal-2025-revisiting}. Prompt-format and answer-symbol binding studies document interface effects \citep{voronov-etal-2024-mind,xue-etal-2024-strengthened}. LoRA supports adaptation tests \citep{hu2022lora}. Cacheable-model studies test rule following with independently encoded inputs \citep{rajput2026cacheable}. Choice2Seq measures correctness retained from complete-answer selection to sequential construction on matched problems \citep{lai2026correctchoicesfailedsequences}. These studies motivate paired evaluations of generalization and presentation sensitivity. Our semantic rule separation and equivalent rewrites measure how adaptation to unfamiliar rule behavior changes across presentations of identical questions with fixed model weights and intended answers.

\paragraph{Information use and internal mechanisms.}
Information-use studies examine how context and representations influence decisions. Performance depends on where relevant context appears \citep{liu-etal-2024-lost}. Probing separates recoverable information from information used to answer \citep{hewitt-liang-2019-designing,belinkov-2022-probing}. Causal abstraction and interchange-intervention training connect representations to high-level variables \citep{NEURIPS2021_4f5c422f,pmlr-v162-geiger22a}. Logit inspection and attention modification examine factual recall under contextual counterfactuals \citep{ortu-etal-2024-competition}. Accuracy gains from supplied values leave their influence on individual answers unresolved. We alter table values and mappings to define alternative targets. Paired transitions toward those targets test directed information use. Source-selected donor replacements separately test whether recovered rule-value information influences the final selected answer.

\begin{takeawaybox}
Choose input interventions that distinguish the behavioral properties behind a score. Separate evidence of transfer, responsiveness, and information use when designing each evaluation.
\end{takeawaybox}

\section{Controlled Rule--State Evaluation}
\label{sec:setup}
\subsection{States, Rules, and Decisions}
A decision instance is $(s,r,t,O)$: a state $s$, an explicitly specified rule $r$, a task family $t$, and an ordered option set $O$ of size $K$. A programmatic oracle returns the semantic answer $y=g_t(r,s)$ and its index in $O$. A model predicts an option index. The semantic answers align option permutations.

Each record has three integer fields $(x,y,z)\in\{-12,\ldots,12\}^3$. Rules follow
\begin{equation}
 f_r(x,y,z)=o_0\bigl(o_1(x,y),o_2(y,z)\bigr),
 \label{eq:grammar}
\end{equation}
where each operator belongs to $\{+,-,\times,\min,\max\}$. Eligible rules depend on all three fields and retain the same three-operator, depth-two structure. Binary decisions test whether $f_r>0$. Ordinal ratings count how many prescribed thresholds the score strictly exceeds, so equality remains in the lower category. Minimum routing selects the record with the lowest rule value. Maximum selection selects the record with the highest value. Record identifiers resolve ties. The option counts are 2 for binary decisions, 3/5/7 for ratings, and 2/4/8 for routing and selection.

\subsection{Semantic Separation and State Sampling}
\label{sec:semantic}
Syntactically different expressions can implement the same decision. We compare numeric, sign, rating, and ranking signatures over the finite record domain, including field permutations. Matching any signature creates an edge between rules. Connected components conservatively keep linked rules in one pool, even when the link is transitive rather than a direct match. The separation covers executable behavior on all records in the enumerated domain.

Replication excludes historically exposed components, then assigns disjoint source and development pools by root-operator quotas and rule characteristics. Reserved components support evaluation of fixed models. Natural sampling draws complete states before labeling. Previously used complete states are excluded across roles, while individual records may recur in different multi-record states. Appendix Algorithm~\ref{alg:construction} summarizes the construction. Appendix~\ref{app:construction-exact} specifies the exact eligibility checks, allocation quotas, and constraints used to construct the paired diagnostic evaluation inputs.

\subsection{Natural Accuracy and Joint Correctness}
Natural accuracy is averaged first within each rule, then equally over rule identities, option counts within each task family, and the four task families. This prevents families with more option-count settings from dominating the pooled macro result.

A diagnostic quartet crosses two rules $r_1,r_2$ with two states $s_1,s_2$:
\begin{equation}
 \begin{aligned}
 Q&=\begin{pmatrix}
 (r_1,s_1)&(r_1,s_2)\\
 (r_2,s_1)&(r_2,s_2)
 \end{pmatrix},\\
 Y_Q&=\begin{pmatrix}a&b\\b&a\end{pmatrix},\ a\ne b.
 \end{aligned}
\end{equation}
Oracle-based eligibility requires the semantic answer to change along either axis. A deterministic predictor invariant to either factor cannot solve the full quartet. Stochastic predictions can succeed by chance, so quartet success remains a behavioral score. We report diagnostic item accuracy and
\begin{equation}
 \begin{aligned}
 \quadacc&=\operatorname{MacroAvg}_{t,K,Q} c(Q),\\
 c(Q)&=\prod_{i=1}^{2}\prod_{j=1}^{2}
 \ind[\hat y(r_i,s_j)=y(r_i,s_j)].
 \end{aligned}
 \label{eq:qacc}
\end{equation}
Diagnostic sampling conditions on quartet eligibility. We compare item accuracy and QAcc on identical diagnostic questions. Natural accuracy retains its original state-sampling distribution, separately from diagnostic-quartet selection criteria.

\lead{The same item score can conceal different rule and state responses.}
The example below shows archived routing predictions for two rules and two states. Both models achieve the same item score while responding differently to the crossed inputs. Appendix~\ref{app:routing-example} gives the inputs and selection rule.

\begin{findingbox}{Same Accuracy, Different Responses}
\small
Rows vary the rule; columns vary the state.
\[
\begin{array}{ccc}
\text{Oracle}&\text{Base}&\text{LoRA}\\[3pt]
\begin{pmatrix}\mathrm{R0}&\mathrm{R1}\\\mathrm{R1}&\mathrm{R0}\end{pmatrix}&
\begin{pmatrix}\mathrm{R1}&\mathrm{R1}\\\mathrm{R1}&\mathrm{R1}\end{pmatrix}&
\begin{pmatrix}\mathrm{R1}&\mathrm{R0}\\\mathrm{R1}&\mathrm{R0}\end{pmatrix}
\end{array}
\]
Both models answer half the items correctly. Base stays invariant; LoRA responds to changes in states alone.
\end{findingbox}

\begin{takeawaybox}
Separate training and evaluation rules by their executable behavior, and isolate complete states across data roles. Cross both input factors to test correct responses to rule and state changes.
\end{takeawaybox}

\section{Behavioral Transfer and Information Use}
\label{sec:experiments}
Three paired diagnostics address the central question by testing transfer, joint correctness, and directed use of the supplied external information.

\begin{figure}[!t]
\begin{keyfindingsbox}
\textbf{Conflict-conditioned transfer:} Gains increase with nearest-source answer agreement. Fully conflicting quartets have low pooled joint success and opposite gain signs across source splits. Equivalent wording changes task-level adaptation gains with the same fixed model weights.\par\smallskip
\textbf{Joint reliability:} Pooled QAcc lies within the primary reference ranges. Maximum-selection departures shrink when calibration preserves ordered rule pairs. Joint-score interpretation depends on the associations retained in the chosen marginal-preserving reassociation reference.\par\smallskip
\textbf{Directed information use:} Jev benefits from correct tables and switches from correct answers to altered-table targets. Paired value and mapping changes distinguish directed information use from an arbitrary response change under paired original and altered-table conditions.
\end{keyfindingsbox}
\end{figure}

\begin{table*}[t]
\centering\small
\begin{minipage}[t]{.53\textwidth}\vspace{0pt}
\centering
\begin{tabular*}{\linewidth}{@{\extracolsep{\fill}}lccc@{}}
\toprule
\rowcolor{tableblue}\multicolumn{4}{c}{\textit{Development rules}}\\
Method & Natural & Diag. item & QAcc\\
\midrule
Controlled Base & 38.67 & 33.81 & 0.52\\
Rank-4 readout & 39.11 & 35.11 & 0.17\\
LoRA & 58.03 & 53.13 & 13.19\\
Jev & 60.44 & 55.34 & 22.05\\
\bottomrule
\end{tabular*}
\end{minipage}\hfill
\begin{minipage}[t]{.44\textwidth}\vspace{0pt}
\centering
\begin{tabular*}{\linewidth}{@{\extracolsep{\fill}}lcc@{}}
\toprule
\rowcolor{tablepeach}\multicolumn{3}{c}{\textit{Reserved rules}}\\
Method & Natural & QAcc\\
\midrule
Controlled Base & 38.06 & 0.00\\
Selected MLP & 42.49 & 1.37\\
LoRA & 57.08 & 9.56\\
\bottomrule
\end{tabular*}
\end{minipage}
\caption{LoRA retains higher natural accuracy on development and reserved rules. Scores are percentages. Development averages two rule splits; reserved evaluation uses batch 16, equal weights for reserved blocks, and MLP readouts selected by source-validation NLL without reserved-rule fitting or any checkpoint selection.}
\label{tab:development}\label{tab:holdout}
\end{table*}

\subsection{Experimental Setup}
\label{sec:protocol}
Controlled comparisons use Qwen3.5-2B-Base, LoRA adaptation, and source-selected frozen readouts \citep{qwen2026qwen35}. Jev is a versioned external decision service \citep{typesafe2026jev}. Appendix~\ref{app:initial-systems} compares additional instruction-tuned and decision systems \citep{decider2026,startlux2026decision}. Appendix~\ref{app:evidence-map} maps each result to its data, readout, precision, batch, and freeze stage.

Each replication cell uses 2,880 source-training questions and their reversed option orders. LoRA adapts decoder linear modules at rank 16 for two epochs, with batch size 16 and optimizer seed 17. Source-natural validation negative log-likelihood selects the checkpoint. Frozen readouts learn residual corrections to candidate logits from the same training questions. Controlled Base and LoRA use BF16 backbone inference with FP32 candidate projection. Independent source-audit questions assess competence after fitting and selection and remain excluded from model fitting and model selection.

\noindent\textbf{Data roles and scoring.} Development evaluates rule transfer, diagnostic quartets, and paired table interventions. Replication crosses two rule splits with three state seeds; the initial comparison varies fit seeds. Their six fixed models share one reserved target pool, later reused for wording and conflict diagnostics. Development comparisons use fixed rank-4 readouts; reserved comparisons use source-selected MLPs. Original reserved scores use batch 16; equivalent-wording comparisons use batch 8.

\begin{table*}[!t]
\centering\small
\setlength{\tabcolsep}{3pt}
\begin{minipage}[t]{.47\textwidth}\vspace{0pt}
\centering
\begin{tabular*}{\linewidth}{@{\extracolsep{\fill}}lrrrr@{}}
\toprule
\rowcolor{tableblue}\multicolumn{5}{c}{Nearest-source agreement groups}\\
Group & Base & LoRA & Gain & QAcc\\
\midrule
\multicolumn{5}{l}{Natural accuracy}\\
Low & 37.06 & 48.53 & +11.48 & ---\\
Middle & 38.31 & 57.71 & +19.40 & ---\\
High & 39.06 & 67.13 & +28.07 & ---\\
\midrule
\multicolumn{5}{l}{Diagnostic item accuracy and QAcc}\\
Low & 34.67 & 41.66 & +6.99 & 6.24\\
Middle & 36.11 & 49.28 & +13.17 & 9.42\\
High & 33.57 & 55.90 & +22.33 & 14.07\\
\bottomrule
\end{tabular*}
\end{minipage}\hfill
\begin{minipage}[t]{.49\textwidth}\vspace{0pt}
\centering
\begin{tabular*}{\linewidth}{@{\extracolsep{\fill}}lrrrr@{}}
\toprule
\rowcolor{tablepeach}\multicolumn{5}{c}{Complete-quartet conflict loads}\\
Load / split & Base & LoRA & Gain & QAcc\\
\midrule
0 / Pooled & 34.91 & 53.49 & +18.58 & 13.99\\
1 / Pooled & 46.49 & 58.19 & +11.70 & 12.32\\
2 / Pooled & 36.62 & 50.76 & +14.14 & 9.37\\
3 / Pooled & 30.96 & 48.12 & +17.16 & 9.03\\
4 / Pooled & 21.44 & 22.83 & +1.39 & 1.42\\
\midrule
\multicolumn{5}{l}{Fully conflicting quartets by source split}\\
4 / Split 1 & 20.47 & 14.37 & -6.10 & 0.00\\
4 / Split 2 & 22.41 & 31.29 & +8.88 & 2.83\\
\bottomrule
\end{tabular*}
\end{minipage}
\caption{Transfer depends on source-answer agreement and conflict. Accuracy and LoRA QAcc are percentages; gains are pp. Base QAcc is zero throughout. Agreement groups rank pairs within task/$K$; load counts conflicting answers per intact quartet. Pooled scores weight source splits equally. Fully conflicting cohorts contain 39 and 32 quartet memberships, covering eight and seven task/$K$ strata. Source splits reuse target questions with potentially different cohort memberships; these finite-cohort scores describe adaptation under source--target conflict.}
\label{tab:transfer-conflict}\label{fig:nearest-conflict}
\end{table*}

\subsection{RQ1: How Do Source Conflict and Wording Shape Transfer?}
\label{sec:transfer}
\lead{LoRA's advantage over the controlled base model persists across development-rule splits.}
Table~\ref{tab:development} retains the LoRA gain across two rule splits. Jev leads overall, with a ranking reversal between splits. The fixed rank-4 readout uses BF16 logit caches, while Controlled Base uses FP32 candidate projection. Appendix Figure~\ref{fig:replication-cells} retains the individual state-seed results for each state seed and source-rule split in the development evaluation.

\lead{LoRA improves reserved-rule performance over its unadapted backbone.}
Table~\ref{tab:holdout} compares fixed LoRA and source-selected MLP readouts with Controlled Base. Models and the LoRA--readout criterion were fixed before reserved-question generation (Appendix~\ref{app:blocks}). LoRA--Base measures adaptation gains. LoRA--readout compares selected training procedures whose source competence remains unmatched (Section~\ref{sec:audits}). Subsequent conflict and wording analyses reuse this reserved benchmark.

\lead{Source-answer agreement separates different transfer demands.}
For each target rule and task/$K$, its nearest source maximizes uniform-domain oracle-answer agreement. We rank target pairs by their mean agreement within each source split and task/$K$, keeping ties together. Table~\ref{tab:transfer-conflict} reports absolute accuracy, gains, and QAcc. Natural gains increase across agreement groups in both splits; pooled QAcc increases, while split-1 QAcc is not monotonic. These relative groups retain all task/$K$ strata, although their absolute agreement ranges can overlap across individual task families.

\lead{Complete source--target conflict exposes split-dependent transfer.}
Conflict load counts differing source and target answers within an intact quartet. At load four, split-1 accuracy falls from 20.47\% to 14.37\%, whereas split-2 accuracy rises from 22.41\% to 31.29\%. Their QAcc is 0.00\% and 2.83\%, respectively. Pooling yields a 1.39 pp item gain and 1.42\% QAcc, concealing the opposing gain signs across source splits. Intermediate conflict loads have no monotonic gain trend. The two source pools reuse target questions, so these contrasts describe source-dependent outcomes on the same fixed target-rule evaluation questions.

On natural conflicting questions, split-1 accuracy changes from 34.49\% to 32.07\% ($-2.43$ pp); split-2 changes from 31.10\% to 43.72\% ($+12.62$ pp). Agreeing subsets improve from 40.83\% to 70.93\% ($+30.09$ pp) and 42.49\% to 70.74\% ($+28.25$ pp), respectively. These conditional scores retain the original evaluation weights. Their gains differ from contributions to the full-set improvement. Individual-item filtering does not define QAcc. Conflict-load analysis instead preserves all four quartet members and reports success against the target oracle across both rules and input states.

Agreement measures similarity over the rule domain; conflict load measures disagreement on the evaluated quartet. Semantic separation alone does not require different answers on every state. These posthoc diagnostics reuse saved predictions and preserve complete quartets and original weights. Appendix~\ref{app:agreement-strata} reports split-level strata, coverage, and conditional intervals; Appendix~\ref{app:native-conflict} reports natural conflict subsets and their gain contributions.

\begin{figure}[t]
\centering
\includegraphics[width=\linewidth]{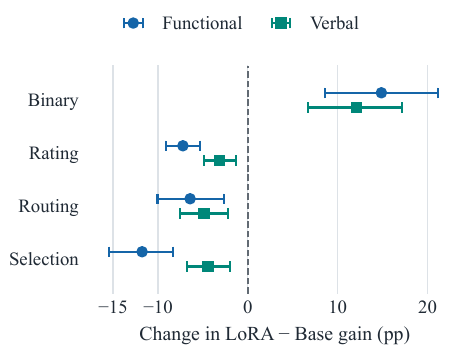}
\caption{Equivalent wording changes task-level adaptation gains in opposite directions. Paired changes in the LoRA--Base advantage use batch 8 and 95\% shared-state intervals. Zero denotes an unchanged advantage on the same semantically equivalent decision questions.}
\label{fig:paraphrase-gain}
\end{figure}

\lead{Comparisons at a fixed batch size reveal opposing task-level responses to equivalent wording.}
We score equivalent rewrites with fixed weights and batch size. Each rewrite parses to the original rule expression. LoRA remains above Base overall, but the functional rewrite reduces its advantage. The verbal rewrite changes little overall because binary gains offset other tasks' declines (Figure~\ref{fig:paraphrase-gain}). Binary gains mainly reflect a larger drop in Base. Elsewhere LoRA declines while Base is stable or improves. These task-level directions persist among items with stable original-text answers across batch sizes. Appendix~\ref{app:round2} reports absolute scores and paired intervals. A fixed subset crosses BF16 and FP32 inference with dynamic and fixed padding. Wording changes continue to alter answers under FP32 with fixed padding, while same-text padding effects shrink under fixed-padding and FP32 model-execution controls (Appendix~\ref{app:numerical-controls}).

\begin{figure}[t]
\centering
\includegraphics[width=\linewidth]{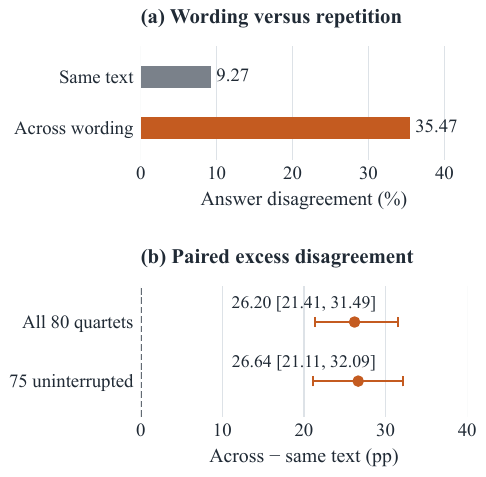}
\caption{Jev changes answers more across equivalent wordings than across identical-text repeats. (a) Disagreement rates average items within splits. (b) The paired excess remains positive after interrupted quartets are excluded. Each wording repeats twice; brackets and whiskers give the 95\% quartet-resampling intervals.}
\label{fig:jev-repeat}
\end{figure}

\lead{Jev changes answers across wording more often than it changes answers across identical-text repetitions.}
Adjacent Jev requests randomize the original and two rewrites, each repeated twice. Cross-wording disagreement exceeds identical-text variability (Figure~\ref{fig:jev-repeat}). The contrast remains positive in all request-slot lag strata and after excluding interrupted quartets. Elapsed-time strata have uneven coverage, including one without same-text pairs. Appendix~\ref{app:adjacent-jev} specifies the paired requests, aggregation weights, and sensitivity checks for the paired, randomized repeated-call comparison.

\begin{takeawaybox}
Measure absolute accuracy, transfer gains, and joint correctness across source-agreement and conflict conditions. Check wording robustness on paired, semantically equivalent questions.
\end{takeawaybox}

\subsection{RQ2: What Does Calibrated Joint Correctness Show?}
\label{sec:joint}

\lead{QAcc measures all-four success; calibration specifies which associations are being tested.}
The primary reference reassociates predictions within cell, task, option-count, quartet-slot, and gold-label strata. It preserves individual prediction marginals while breaking the observed quartet associations. Pooled LoRA and Jev QAcc lie within these reference ranges, while maximum-selection scores exceed them (Table~\ref{tab:calibration}). Retaining ordered rule pairs brings maximum-selection reference means closer to observed QAcc. The state-vector reference additionally retains both predictions on a state together, preserving their original within-state response associations (Appendix~\ref{app:pairing-controls}).

Preserving rule pairs yields pooled reference means of 13.33\% for LoRA and 21.69\% for Jev. Adding state-vector preservation gives 12.67\% and 21.61\%, respectively. Observed QAcc remains 13.19\% and 22.05\% across these references. A departure concerns associations removed by that specific reference; it does not identify an internal reasoning mechanism. Falling within a reference range leaves the observed all-four success unchanged. Finer conditioning can restrict movable predictions, and the state-vector reference does not match latent state difficulty. These controls characterize the constructed pairs. Reliability under independently sampled second states remains untested.

\begin{table}[t]
\centering\small
\setlength{\tabcolsep}{4pt}
\begin{tabular}{lrrr}
\toprule
\rowcolor{tableblue}
Method & QAcc & Ref. mean & \shortstack{95\% reference\\range}\\
\midrule
Base & 0.52 & 1.09 & [0.00, 2.60]\\
Rank-4 readout & 0.17 & 1.05 & [0.17, 2.09]\\
LoRA & 13.19 & 12.21 & [10.07, 14.41]\\
Jev & 22.05 & 21.29 & [19.79, 22.92]\\
\midrule
\multicolumn{4}{l}{Maximum selection only}\\
LoRA & 13.19 & 7.60 & [4.17, 10.42]\\
Jev & 13.19 & 8.75 & [6.25, 11.81]\\
\bottomrule
\end{tabular}
\caption{Pooled LoRA and Jev QAcc lie within the marginal-preserving reference ranges, while maximum-selection scores exceed their ranges. Scores are percentages. Central 95\% ranges describe 1,000 stratified reassociations that preserve prediction marginals; they summarize the reference distribution rather than confidence intervals for performance on newly sampled rules.}
\label{tab:calibration}
\end{table}

\lead{Predictors that ignore rules or states can score on individual items but cannot solve full quartets.}
Source-label-majority, rule-blind, and state-blind controls demonstrate the gap between item and quartet success. Random choices supply a chance reference. Appendix~\ref{app:calibration} defines each control and the grouped quartet uncertainty estimates.

\lead{Aggregate rankings conceal distinct task profiles.}
Appendix Figure~\ref{fig:p4-task-profile} shows Jev\textquotesingle{}s strongest joint performance on binary decisions. LoRA leads on ordinal rating and minimum routing and ties Jev on maximum selection. Task-level QAcc distinguishes joint reliability on these rule families.

\lead{Both rule changes and state changes can fail to trigger the required response.}
LoRA reduces invariant predictions along both axes relative to Controlled Base. Remaining errors include unchanged answers and incorrect changes. Appendix~\ref{app:p1-structure} separates invariant predictions, incorrect changes, and partial success by counting correct answers within each complete crossed quartet of rules and states.

\begin{takeawaybox}
Calibrate joint correctness against references with explicitly stated preserved associations. Interpret each departure relative to the associations removed, and report all-four success separately from its defined reference comparison.
\end{takeawaybox}

\subsection{RQ3: How Do Answers Respond to External Values?}
\label{sec:information}
D2 supplies every rule\textquotesingle{}s value for every record. D3 marks the same entries unavailable. Both retain the state, question, options, answer, and supplementary-table instructions. Using a common all-rule table requires identifying the requested rule. Appendix~\ref{app:binding} defines the conditions, and Appendix~\ref{app:jev-native} provides the full native Jev requests.

\begin{figure*}[t]
\centering
\includegraphics[width=\linewidth]{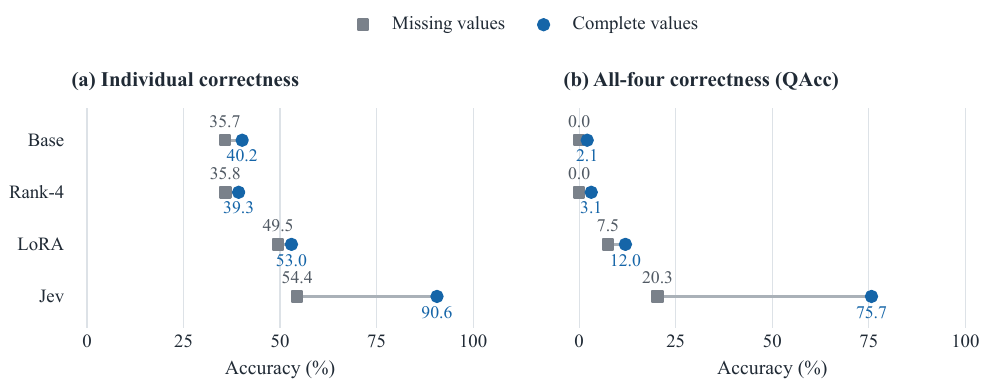}
\caption{Complete value tables benefit Jev more than the local predictors on these paired inputs. Lines connect missing-value (D3) and complete-value (D2) scores. Endpoint labels use a shared 0--100\% accuracy scale.}
\label{fig:values}
\end{figure*}

\lead{The same external values benefit the tested systems to different degrees.}
Complete values improve Jev consistently across replication cells. Local models gain less. LoRA\textquotesingle{}s effects vary across cells, revealing differences in how the same supplied values affect full-quartet success (Figure~\ref{fig:values}).

\lead{Jev's answers shift toward targets implied by altered values and mappings.}
We mismatch formula rows or record columns, or change target-formula values to imply another answer. The original oracle stays fixed. Among conflicting questions answered correctly with the correct table, Jev usually switches to the altered-table target (Table~\ref{tab:answer-transfer}). Both mapping mismatches preserve the value multiset and table character length. The pattern persists under task-balanced and equal-cell weighting (Appendix~\ref{app:weighting}). Appendix~\ref{app:review-controls} reports all tested systems.

A transition pairs an original correct answer with the new table target. Let $T$ count these transitions, $N$ all applicable conflicts, and $B$ those initially answered correctly. Table~\ref{tab:answer-transfer} reports $T/N$ and $T/B$, using the pooled integer counts across all applicable evaluated task families and option-count strata.

\begin{table}[t]
\centering\small
\setlength{\tabcolsep}{4pt}
\begin{tabular}{@{}lrr@{}}
\toprule
\rowcolor{tableblue}
Jev intervention & \shortstack{Overall\\$T/N$ (\%)} & \shortstack{Initially correct\\$T/B$ (\%)}\\
\midrule
Formula rows & \shortstack{169/220\\(76.82)} & \shortstack{169/191\\(88.48)}\\
Record columns & \shortstack{173/190\\(91.05)} & \shortstack{173/177\\(97.74)}\\
Target values & \shortstack{243/320\\(75.94)} & \shortstack{243/282\\(86.17)}\\
\bottomrule
\end{tabular}
\caption{Jev usually switches to the altered-table target. Each cell gives transitions over its denominator, followed by the percentage. Overall rates use all conflicts; conditional rates use the initially correct conflicts.}
\label{tab:answer-transfer}
\end{table}

\lead{Even with the correct answer supplied, performance depends on context and instruction wording.}
Crossing context length with answer-binding instructions changes how systems use supplied answers. Local models benefit from direct instructions, while Jev performs best with legacy instructions and minimal context (Appendix Figure~\ref{fig:binding-controls}).

\begin{takeawaybox}
Edit supplied values and their mappings to define alternative decision targets. Measure transitions toward those targets and separately test whether the answer interface correctly binds supplied information to the selected answer.
\end{takeawaybox}

\section{Source Competence and Internal Interventions}
\label{sec:audits}
\lead{Independent source auditing leaves source competence unmatched.}
The qualification gate requires natural accuracy of 90\% and QAcc of 75\% on training and independent source-audit inputs. Base, selected readouts, and LoRA fail this gate in all six replication cells. Common-audit natural accuracy / QAcc is 36.31\% / 0.00\% for Base, 45.70\% / 1.39\% for selected readouts, and 75.43\% / 35.94\% for LoRA. Validation-based early-checkpoint matching succeeds in two cells, but independent audit matching succeeds in none (Appendix~\ref{app:source-match-review}). LoRA--readout differences therefore compare the attained outcomes of the selected training procedures, with source competence still unmatched. They do not estimate a transfer advantage after controlling for attained source competence.

Source-selected internal donor replacements yield no stable rule-specific improvement over controls. Appendix~\ref{app:mechanisms} reports probe results, intervention norms, and explicit paired positive controls.

\begin{takeawaybox}
Check source competence on independent inputs before attributing differences to transfer. Validate internal interventions with restoration and positive controls before interpreting the recovered rule information as decision-relevant.
\end{takeawaybox}

\section{Conclusion}
We asked which input differences guide decision models' answers. Controlled tests distinguish conflict-conditioned transfer, all-four reliability, and directed information use. Adaptation gains increase with nearest-source answer agreement, while complete conflict yields opposite gain signs across source splits and low pooled QAcc. Equivalent wording changes task-level gains, and calibration conclusions depend on the associations retained by the reference. Paired value and mapping interventions identify directed use of supplied information. Evaluation and method design should distinguish source-consistent success from correct decisions under conflict and preserve correct answers across equivalent wording. Report joint success with explicitly specified prediction-reassociation calibration conditions. Method comparisons should report attained source competence alongside target scores. Joint reliability remains a design objective even when individual decision accuracy improves under unfamiliar rules or supplied information.

\clearpage
\section*{Limitations}
\lead{The findings concern the tested rule families and evaluation conditions.}
The core grammar uses fixed-depth arithmetic rules on three bounded integer fields. Semantic separation is established within this finite domain. Boolean and finite-state extensions test additional rule formats rather than broad natural-language policy coverage. Fully conflicting cohorts contain 39 and 32 quartet memberships with different task coverage. Their opposing gains describe source-dependent transfer within the observed task coverage under complete conflict.

LoRA--readout comparisons estimate outcomes of the selected training procedures, because independent source-audit competence remains unmatched. The larger-model comparison also changes system identity, so the observed effects cannot be attributed to parameter count alone.

QAcc describes joint success on the constructed rule--state quartets. Calibration departures concern the associations removed by each reassociation reference. Small or constant-prediction strata restrict the available reassociations. Multi-record second states use swapped winning-record fields, so the state-vector reference does not establish reliability under independently sampled pairs of input states.

Wording results compare the original rendering with two equivalent templates. Jev's adjacent repeats estimate variability within the observed service window, rather than performance across service versions or observation periods. Table interventions establish directed use of supplied information in answers; they do not identify an internal decision mechanism. Model comparisons concern these controlled diagnostics, rather than comparative detection rates across existing behavioral tests.

\section*{Ethics Statement}
We use synthetic states and oracle labels, with no human participants or personal records. Remote services receive only synthetic inputs. Benchmark performance does not establish suitability for consequential decisions. Generative AI tools assisted implementation, literature organization, and writing. Experimental records, failure evidence, and the actual scope of each analysis are retained.

\bibliography{references}

\clearpage
\appendix

\section{Task Construction and Evaluation Protocol}
\label{app:protocol}
\subsection{Executable Rules and Task Semantics}
The rule grammar in Equation~\ref{eq:grammar} contains three operator slots drawn from five operators, giving $5^3=125$ syntactic candidates. Each expression is evaluated over all $25^3=15{,}625$ integer records in $\{-12,\ldots,12\}^3$. Eligibility requires effective dependence on all three fields and excludes expressions whose audited task signatures match tested lower-complexity expressions. This leaves 104 candidates. Numeric, binary, ordinal, and full weak-order signatures, including ties and field permutations, link equivalent candidates into 77 connected components. Using components rather than pairwise exclusions also prevents a chain of equivalences from spanning data pools. The audit establishes behavioral separation over the enumerated integer domain. Symbolic equivalence on real inputs lies outside this finite integer-domain test.

For one-record tasks, let $v=f_r(x,y,z)$. Binary decisions return \texttt{yes} exactly when $v>0$, and \texttt{no} otherwise. Ordinal ratings with $K\in\{3,5,7\}$ return
\begin{equation}
 y=\sum_{i=1}^{K-1}\ind[v>10(i-\lfloor K/2\rfloor)].
\end{equation}
The threshold lists are $[0,10]$, $[-10,0,10,20]$, and $[-20,-10,0,10,20,30]$ for $K=3,5,7$, respectively. Equality to a threshold does not increment the rating. Multiclass routing uses $K\in\{2,4,8\}$ records and selects the smallest score. Structured selection uses the same option counts and selects the largest score. Both resolve ties by the smallest numeric record ID, independently of candidate order. There are ten task/$K$ strata, with equal macro weight assigned to each of the four task families.

\subsection{Rule Pools and Complete-State Isolation}
Before replication, a historical ledger contains 35,940 complete states and 40 rule ASTs. Historical exposure excludes 19 of the 77 components. Four disjoint pools of eight components use 32 of the remaining 58, forming two source/development splits and leaving 26 components in reserve. Each eight-rule pool has root-operator counts of one addition, two subtractions, two multiplications, one minimum, and two maxima. Pool construction also matches recorded operator-composition and binary-answer characteristics. The chosen pools pass a separate ordinal overlap check at each $K$. Rule split seeds are 20261101 and 20261102. State seeds within each split are 101, 202, and 303.

Natural sampling first draws complete states with integer coordinates uniform on the finite domain, then evaluates every rule in the relevant pool. It does not filter states by answer, class balance, margin, or tie. Rejection excludes complete states already assigned to historical data or another role. A complete-state key is the ordered sequence of all records' $(x,y,z)$ triples. This isolates entire states. An individual record can occur in two otherwise distinct multi-record states. Source training, source validation, development, and their natural/diagnostic subsets have disjoint complete-state keys. Replication cells also have disjoint complete states, while questions for different rules within a natural role intentionally share complete states.

\subsection{Semantic Reconstruction and Similarity Audit}
Exhaustive signatures over the finite domain and all six field permutations show no cross-pool equivalence, including complete weak orders for record selection. Non-equivalent rules may still agree frequently. Appendix~\ref{app:round2} measures nearest-source oracle agreement under the original and best-permuted fields, separating uniform-domain similarity from accepted natural states. The source and reserved rule allocations remain fixed throughout these posthoc, paired source--target agreement analyses.

\subsection{Exact Eligibility and Pool Construction}
\begin{algorithm}[t]
\small
\caption{Construct rule pools and evaluations}
\label{alg:construction}
\begin{algorithmic}[1]
\Require Rules $G$, simpler rules $L$, record domain $D$, exposed components $H$, used states $U_0$, seeds
\Ensure Rule pools $P$, natural sets $N$, quartets $Q$
\For{$r\in G\cup L$, signature type $k$}
  \State $\sigma_k(r)\gets\min_{\pi\in S_3}\mathrm{Hash}(\mathrm{Sig}_k(r\circ\pi,D))$
\EndFor
\State $R\gets\{r\in G:\mathrm{Eligible}(r,L,\sigma)\}$
\State $E\gets\{(r,r')\in R^2:\exists k,\ \sigma_k(r)=\sigma_k(r')\}$
\State $C\gets\mathrm{ConnectedComponents}(R,E)$
\State $P\gets\mathrm{SelectPools}(C\setminus H,\mathrm{quotas,seeds})$
\State $N,Q\gets\varnothing,\varnothing$, $U\gets U_0$
\For{pool $p$, data role, and valid task/$K$ pair $(t,K)$}
  \State $S\gets\mathrm{SampleStates}(\mathcal S_{t,K}(D)\setminus U)$
  \State $N\gets N\cup\{(r,s,g_t(r,s)):r\in p,s\in S\}$
  \State $U\gets U\cup S$
  \State $Q'\gets\mathrm{SampleQuartets}(p,\mathcal S_{t,K}(D)\setminus U)$
  \Statex\hspace{\algorithmicindent}subject to $y_{00}=y_{11}\ne y_{01}=y_{10}$,
  \Statex\hspace{\algorithmicindent}where $y_{ij}=g_t(r_i,s_j)$
  \State $Q\gets Q\cup Q'$, $U\gets U\cup\mathrm{States}(Q')$
\EndFor
\State \Return $\mathrm{Freeze}(P,N,Q,\mathrm{seeds,hashes})$
\end{algorithmic}
\end{algorithm}
Here $\mathcal S_{t,K}(D)$ contains complete task states, including ordered records. Sampling uses the fixed seeds, rule quotas, and per-task record-count strata.

\label{app:construction-exact}
Let $F=\{x,y,z\}$ and $O=\{+,-,\times,\min,\max\}$. The simpler-expression set starts with $F$ and $U=\{o(a,b):o\in O,\ a,b\in F\}$, allowing repeated leaves. It also includes $|a|$, $o(u,a)$, $o(a,u)$, $|u|$, $o(|a|,b)$, and $o(a,|b|)$ for $a,b\in F$ and $u\in U$. We deduplicate expression trees by sorting commutative children and canonicalizing field names under each of the six possible integer-field coordinate permutations.

For each expression, the numeric signature contains all integer values over the domain. The binary signature contains the signs $f>0$. The ordinal signature jointly stacks the answers for $K\in\{3,5,7\}$. The ranking signature gives the rank of each distinct value and retains ties. This complete weak order records every less-than, equal-to, and greater-than relation, rather than just the winning record. Each canonical signature is the minimum hash across the six coordinate permutations. An expression is rejected if any corresponding canonical signature agrees with a simpler expression, if any field never changes the output, or if its binary answer is constant. Graph edges join expressions when \emph{any one} of the four canonical signatures agrees, and connected components take the transitive closure. Per-$K$ ordinal overlap is checked separately across pools. A graph edge therefore requires one matching signature, rather than a simultaneous match of behavior across all four task-specific signatures.

After excluding exposed components, each split uses root quotas $(1,2,2,1,2)$ for $(+, -,\times,\min,\max)$. Within each root class, a seeded choice supplies one representative, and its partner minimizes the $L_1$ difference in operator-count vectors plus ten times the absolute difference in binary-positive frequency. Formula strings break distance ties, and a seeded coin assigns the pair to source or development. Each pool must permit four disjoint rule pairs for every task/$K$. Binary and rating pairs need at least 16 unused states in each reversal bucket. Multi-record pairs are screened by 256 deterministic seeded probes before pair matching. The procedure allows at most 10,000 attempts. Accepted components are removed before selecting the next split. The 26 remaining components retain root counts of at least $(4,4,4,2,2)$ and form the held-out reserve.

\subsection{Counterbalanced Diagnostic Construction}
Binary and ordinal generation samples unused states from opposite-answer strata. For each task/$K$/role, the generator traverses the frozen disjoint rule pairs in order and constructs two quartets per pair. For multi-record tasks it draws $s_0$, obtains the two winning record IDs $a,b$, and rejects $a=b$. It copies $s_0$ to $s_1$ and swaps the $(x,y,z)$ fields of records $a,b$, retaining their IDs. Both rules are evaluated again. Acceptance requires the checkerboard labels, two different complete states, and neither state in the used-state set. Accepted states enter that set. A quartet gets at most 20,000 attempts, after which construction fails explicitly. The four items are emitted in state-major, then rule-major order. Construction uses the oracle answers only.

Each source-training question also receives one reversed candidate order, preserving semantic answers and record-ID tie-breaking. Reversed quartets are separate four-item groups, not eight-item success events. Cross-rendering comparisons align semantic answers. Diagnostic sampling conditions on quartet eligibility. We separately report diagnostic scores and natural accuracy on their respective sets of sampled natural and diagnostic inputs.

\subsection{Evaluated Quartet and Local Decision Prompt}
\label{app:local-input}
\lead{The evaluated quartet changes its answer along both input axes.}
We take the first binary development quartet in replication split 1, state seed 101, pair 0, quartet 0. Its two component representatives are $r_0=(x\times y)+\max(y,z)$ and $r_1=(x\times y)-(y-z)$. The stored states and recomputed exact scores are shown in Table~\ref{tab:archived-quartet}. Dataset order selects this example before model predictions are examined, preserving its role as an illustration of the paired, crossed rule and state construction.

\begin{table}[t]
\centering\small
\begin{tabular}{lcccc}
\toprule
\rowcolor{tableblue}
State $(x,y,z)$ & $r_0$ & Answer & $r_1$ & Answer\\
\midrule
$(0,-8,0)$ & $0$ & no / A & $8$ & yes / B\\
$(0,6,-10)$ & $6$ & yes / B & $-16$ & no / A\\
\bottomrule
\end{tabular}
\caption{Changing either the rule or the state reverses the binary answer in this evaluated quartet. Exact values determine the gold answers; zero has the answer no.}
\label{tab:archived-quartet}
\end{table}

The template below reproduces the local candidate-scoring input. Blue braces mark substituted fields; they are notation, not literal prompt characters. The evaluated instance uses \texttt{state} = \texttt{\{"facts": \{"id": "R0", "x": 0, "y": -8, "z": 0\}\}}. Its formula is \texttt{((x * y) + max(y, z))}, and its options are \texttt{A) no} and \texttt{B) yes}. Substitution recovers the original first-item prompt. The gold A (no) is evaluation-only. The displayed field and option breaks are literal newlines. Additional line wrapping serves presentation and does not change the evaluated local candidate-scoring prompt.

\begin{promptbox}{Binary decision: local candidate scoring}
\footnotesize\ttfamily\raggedright
State: \promptfield{state}\par
\smallskip
Question: Using facts, is \promptfield{formula} strictly greater than 0? Select no or yes. Only x, y, z participate. Ignore other fields and notes.\par
\smallskip
Options:\par
\promptfield{letter}) \promptfield{semantic option}\par
\promptfield{letter}) \promptfield{semantic option}\par
Answer:
\end{promptbox}

\subsection{Routing Example}
\label{app:routing-example}
The main-text matrix uses $r_0=\min(x,y)-(y+z)$ and $r_1=\min(\min(x,y),y-z)$. State $s_0$ contains R0=$(-9,1,8)$ and R1=$(5,-5,7)$, and $s_1$ exchanges their fields. The task selects the minimum-scoring record. The example is the first lexicographic archived case with equal correct counts but different positions of the correct model answers.

\subsection{Result Configurations and Data Roles}
\label{app:evidence-map}
The initial system comparison uses the initial batch and averages LoRA over fit seeds 17, 29, and 43. Replication uses two rule splits with three state seeds per split and optimizer seed 17. Its fixed rank-4 readout retains BF16 logit caches, whereas Controlled Base uses FP32 candidate projection.

Reserved evaluation freezes six source-selected MLP readouts and six LoRA models before generating questions. The LoRA--readout margin criterion is prospective, and Table~\ref{tab:holdout} uses batch 16 with equal rule-block weights. Later wording analyses use the inspected reserved benchmark and rescore the original and both rewrites at batch 8. This explains the batch-dependent differences between Table~\ref{tab:holdout} and the wording tables. The third split repeats development on previously exposed components.

Quartet calibration reuses replication predictions. Table interventions use paired development questions, and adjacent Jev repeats use fixed development quartets. These later analyses retain model and input identities while introducing their stated conditions. Natural-score intervals group shared complete states, diagnostic intervals group complete quartets, and reserved analyses retain rule-block weights. Sections~\ref{app:calibration} and~\ref{app:adjacent-jev} define uncertainty conditional on the fixed model checkpoints and the paired evaluation questions and the dependencies among shared complete states or quartets.

\subsection{Data Roles and Permitted Selection}
The initial comparison gives each system 1,920 original-task questions: 640 natural and 320 diagnostic questions in each of source and development. Auxiliary conditions reuse the 320 development diagnostic indices. Three initial LoRA optimizer seeds share rules and states. Replication instead fixes optimizer seed 17 and repeats each of two rule allocations with state seeds 101/202/303. Table~\ref{tab:data-roles} specifies all roles. Each cell has 960 complete states and 7,680 rendered rows before audits and auxiliary inputs. Candidate reversal reuses states.

\begin{table*}[t]
\centering\small
\begin{tabularx}{\textwidth}{p{3.8cm}rrX}
\toprule
\rowcolor{tableblue}
Role per replication cell & Questions & States & Permitted use\\
\midrule
Source natural training & 2,560 & 320 & Gradient fitting. Source-training capability report.\\
Source diagnostic training & 320 & 160 & Gradient fitting. Source-training quartet report.\\
Reversed source renderings & 2,880 & 0 new & Same labels after semantic option remapping. Fitting augmentation.\\
Source natural validation & 640 & 80 & Checkpoint, architecture, feature, and regularization selection where those choices are in the protocol.\\
Source diagnostic validation & 320 & 160 & Joint-response reporting and earlier operational qualification. Not the checkpoint-selection objective.\\
Development natural & 640 & 80 & New-rule natural evaluation.\\
Development diagnostic & 320 & 160 & New-rule item/quartet evaluation and matched auxiliary conditions.\\
\midrule
Later source natural audit & 640 & 80 & Qualification after richer-readout selection is locked. No reselection.\\
Later source diagnostic audit & 320 & 160 & Quartet qualification after richer-readout selection is locked. No reselection.\\
\bottomrule
\end{tabularx}
\caption{Training, selection, and evaluation use distinct data roles in each replication cell. Reversed renderings augment training without adding complete states; audit data remain excluded from all model-selection decisions.}
\label{tab:data-roles}
\end{table*}

Source competence is audited on 960 questions per cell using fresh states and the original source rules. These audits measure competence after source-natural selection and do not choose the feature, architecture, regularization, or seed. The six seed-17 selected readouts and six replication LoRA checkpoints are frozen before generating questions for the 26 reserved components. Its coverage and predeclared decision rule appear in Appendix~\ref{app:blocks}.

\subsection{Macro Metrics and Paired Comparisons}
Let $\mathcal T$ be the four task families, $\mathcal K_t$ their option counts, $\mathcal R_{tK}$ the rules present in a natural stratum, and $I_{tKr}$ its item indices. For per-item score $a_i$, define
\begin{align}
 \mathcal M(a)={}&\frac{1}{|\mathcal T|}\sum_{t\in\mathcal T}
 \frac{1}{|\mathcal K_t|}\sum_{K\in\mathcal K_t}\bar a_{tK},\\
 \bar a_{tK}={}&\frac{1}{|\mathcal R_{tK}|}\sum_{r\in\mathcal R_{tK}}
 \frac{1}{|I_{tKr}|}\sum_{i\in I_{tKr}}a_i.
 \label{eq:macro-detail}
\end{align}
Natural accuracy uses $a_i=\ind[\hat y_i=y_i]$. Source selection uses $a_i=-\log p_i(y_i)$, the candidate-normalized negative log-likelihood. This rule-level average prevents an imbalance in rule counts from changing a task's weight. Training cross-entropy instead weights task families and then option counts equally over the available training rows. The objective balances task and option-count strata without balancing the observed semantic answer classes.

For diagnostic quartets $\mathcal Q_{tK}$, let $c_{qj}$ indicate correctness at position $j$ and $C_q=\sum_{j=1}^4c_{qj}$. The within-stratum item score is $\mathbb E_q[C_q/4]$ and the all-four score is $\mathbb E_q[\ind(C_q=4)]$. Both are averaged over $K$ and task as in Equation~\ref{eq:qacc}. We also report the distribution $\Pr(C_q=c)$ for $c=0,\ldots,4$. These macro distributions need not yield integer group counts when multiplied by the number of quartets. Paired rule- and state-switch outcomes distinguish both answers correct, an unchanged semantic prediction, and a changed semantic prediction with an incorrect answer in either item.

Replication means average the three state seeds within each rule split and then the two splits equally. D2--D3 contrasts are paired within each cell before this aggregation. Reserved-rule comparisons first aggregate within a rule block, then over the six fixed model pairs, then over the 13 blocks equally. Diagnostic reserved-rule scores cover feasible strata only. Missing quartets are disclosed rather than imputed. Shared rules, states, model fits, and quartet members are retained as dependencies. The aggregation retains these shared dependencies. Quartet success is measured on the crossed questions rather than inferred from the fourth power of the separately measured natural item accuracy.

\section{Model Interfaces, Training, and Readout Selection}
\label{app:training}
\subsection{Input Rendering and Candidate Scoring}
Controlled Base, LoRA, and the later frozen readouts use the prose renderer exemplified in Appendix~\ref{app:local-input}. The state, explicit formula, task instructions, and semantic candidate text are available in the input. Gold labels remain withheld.

Candidate letters are checked to be distinct single nonspecial tokens after the answer prefix. In the controlled path, the BF16 text backbone produces the final-normalized hidden representation $h$ and its legal candidate logits are
\begin{equation}
 z_k=E_k^\top h,\qquad
 p_k=\frac{\exp z_k}{\sum_{j=1}^{K}\exp z_j},
\end{equation}
with the candidate projection evaluated in FP32 using the original output rows $E_k$. Semantic prediction is the candidate argmax. Training optimizes this restricted candidate loss rather than a free-text generation objective. The original embeddings and candidate rows stay frozen during LoRA adaptation. Inputs are encoded without an explicit length-truncation cap and left-padded to the longest prompt in each batch. Appendix~\ref{app:numerical-controls} separately tests fixed padding and FP32 model inference, distinct from the FP32 candidate projection used here.

External systems retain their recorded interfaces: Decider uses its decision prompt, StartLux its native interface, and Jev the native TypeSafe structured request. Such comparisons characterize deployed system configurations and do not isolate a common architecture or training recipe. Table~\ref{tab:identities} records the checkpoint and service identifiers.

\subsection{LoRA Adaptation}
LoRA is attached to all linear modules of the text decoder in the common Qwen Base checkpoint, with rank 16, $\alpha=32$, scale $\alpha/r=2$, dropout 0.05, and no trained bias. Only adapter tensors within the language decoder are trainable, totaling 16,819,200 parameters. Each replication cell starts from the same unadapted checkpoint. Training uses AdamW at learning rate $10^{-4}$, weight decay $10^{-4}$, gradient-norm clipping at 1, batch size 16, and no scheduler. Non-reentrant gradient checkpointing is enabled. Two epochs over 5,760 rows yield 720 optimizer updates per cell. Source-natural macro NLL chooses between the epoch-1 and epoch-2 checkpoints, with any exact tie resolved in favor of the earlier of the two saved epoch checkpoints.

The data include original and reversed candidate orders, and task/$K$ weighting is applied to candidate cross-entropy. The procedure does not use oracle intermediate-value labels, development examples, or audit scores for training or selection. The six replication fits share optimizer seed 17. The initial fits additionally use seeds 29 and 43 on their common dataset. These two uses of seed variation have different scientific interpretations.

\subsection{Frozen Residual Readouts}
All learned frozen readouts add a correction to the base logits rather than replacing them:
\begin{equation}
 z'=z_{\mathrm{base}}+g_\theta(\tilde h),\qquad
 \tilde h_j=\frac{h_j-\mu_j}{\max(\sigma_j,10^{-6})}.
 \label{eq:readout-detail}
\end{equation}
Population means and standard deviations use source-training rows only. Scores beyond the legal $K$ candidates are masked. The fixed rank-4 readout uses $g_\theta(\tilde h)=(\tilde h V)(EU)^\top$, with $U,V\in\mathbb R^{2048\times4}$, fixed candidate rows $E$, and zero-initialized $U$. It trains with learning rate 0.003 and L2 strength $10^{-4}$ for 800 updates, evaluating source-natural NLL every 40 steps. Replication reuses this configuration.

Readout selection compares three feature sets. The final-normalized final-position vector has 2,048 dimensions. Concatenating final-position outputs at layers 8 and 16 with the final-normalized output gives 6,144 dimensions. Final-layer means over the state, question, and option regions also give 6,144 dimensions. The question region includes the rule text. These rendered-input features omit the numeric values computed by the oracle.

Each feature set is paired with a linear correction or a GELU MLP with 128 hidden units and eight outputs. L2 strength is selected from $\{10^{-4},10^{-2}\}$, and the output layer starts at zero. Both families minimize
\begin{equation}
 \mathcal L(\theta)=\sum_{i\in\mathcal D_{\mathrm{train}}}
 w_i\bigl[-\log p_\theta(y_i\mid i)\bigr]
 +\frac{\lambda}{2}\|\theta\|_2^2,
\end{equation}
where $w_i$ gives equal mass to tasks and then option counts. L2 covers trained weights and biases. Linear corrections use FP64 L-BFGS with strong-Wolfe line search, at most 40 chunks of 20 iterations, and history size 30. The stopping threshold is a maximum absolute gradient of $10^{-6}$, with source validation after each chunk. MLPs use FP32 full-batch Adam at learning rate 0.003 for 800 updates, with gradient clipping at 1. We evaluate source-validation NLL after every 40 optimizer updates for source-based checkpoint selection using the same candidate-normalized decision-scoring rule.

Source-natural validation NLL selects the configuration with seed 17. Exact ties favor fewer parameters, then the earlier checkpoint, then the fixed configuration order. Seeds 29 and 43 refit the selected configuration without reselection. Every cell selects MLP128 with $\lambda=0.01$. Split 1/state 202 uses multilayer features with 787,592 parameters. The other cells use final-position features with 263,304 parameters. Reserved-rule evaluation uses the six seed-17 selected checkpoints. LoRA and readouts share training examples but differ in parameter count, optimization, and computation.

The correct-count figure also reports rank-4 readouts fitted with cross-entropy, true-pair regularization, or random-pair regularization. These variants use 800 full-batch Adam updates at 0.003, with L2 selected from $\{10^{-4},10^{-2}\}$. Source-natural NLL selects checkpoints every 40 updates. True pairs share a state but use different rules and gold answers. Random pairs reassign the second member within task/$K$/directed-label-pair strata, requiring different states. Strata without a valid reassignment are omitted from both paired variants. For labels $(a,b)$, define
\begin{equation}
 d_{ij}=(z'_{i,a}-z'_{i,b})-(z'_{j,a}-z'_{j,b}).
\end{equation}
The added task/$K$-weighted loss is $0.25\,\operatorname{softplus}(1-d_{ij})$. Both paired variants retain identical cross-entropy training rows and the same source-natural validation selection criterion.

\subsection{Precision and Interface Qualification}
\label{app:precision}
\begin{table*}[t]
\centering
\footnotesize
\begin{tabular}{ll}
\toprule
\rowcolor{tableblue}
Model repository & Frozen revision / response version\\
\midrule
\texttt{Qwen/Qwen3.5-2B-Base} & \texttt{b1485b2fa6dfa1287294f269f5fb618e03d52d7c}\\
\texttt{Qwen/Qwen3.5-2B} & \texttt{15852e8c16360a2fea060d615a32b45270f8a8fc}\\
\texttt{Mapika/decider-2b} & \texttt{533964dae8be954c5b5e19fa4948e48408094c1e}\\
\texttt{startlux-models/StartLux-Decision-2B} & \texttt{3e2e456409a7fe44be69eee231566f7830872ea2}\\
TypeSafe Jev (request alias: \texttt{jev-latest}) & \texttt{jev-1.13.0} (successful responses)\\
\bottomrule
\end{tabular}
\caption{Fixed checkpoint revisions and the recorded service version identify the evaluated systems. Jev's recorded version identifies the service response, while its internal checkpoint and parameter count remain undisclosed.}
\label{tab:identities}
\end{table*}
The fixed rank-4 replication cache uses BF16 logits from the full model output and FP32 hidden features. Its BF16 Base reference is distinct from Controlled Base. The later richer readouts and reserved-rule evaluation use the controlled FP32 candidate-letter projection shared with Base and LoRA. The rank-4 versus controlled-Base contrast therefore combines readout differences with a change in the numerical precision of candidate-logit projection.

StartLux quality scores use FP32 inference. The probability-equivalence check passes with a maximum difference of 0.000800 against tolerance 0.01. The check verifies probability agreement for the scoring interface within the prescribed tolerance.

Jev's successful responses all identify version \texttt{jev-1.13.0}. In initial repeat controls, semantic agreement is 90.625\% across 32 repeated requests, with maximum probability change 0.12. The 43 eligible option-reversal comparisons have 93.02\% semantic agreement. Elapsed time and service stochasticity are not isolated by that comparison. We therefore do not treat the service as deterministic. Source, development, and reserved labels specify the evaluation allocation; they make no assertion about an external model's pretraining data.

\section{Detailed Behavioral Results}
\subsection{Initial System Comparison}
\label{app:initial-systems}
\lead{Higher item accuracy coexists with weak joint correctness across decision systems.}
Table~\ref{tab:initial-system-overview} compares released systems and controlled adaptation. LoRA improves the controlled backbone, while decision systems show different source and development strengths. Their interfaces and training histories define distinct system configurations.

The initial comparison averages LoRA over three fit seeds. Source scores use validation inputs. Natural and diagnostic scores use separately sampled questions. Native BF16 and controlled FP32 projection paths are identified separately for each system in the result table and Appendix~\ref{app:training}.

\begin{table*}[t]
\centering\small
\setlength{\tabcolsep}{4pt}
\begin{tabular*}{\textwidth}{@{\extracolsep{\fill}}lcccc@{}}
\toprule
\rowcolor{tableblue}
\multicolumn{1}{c}{System} & \multicolumn{1}{c}{Source rules} & \multicolumn{3}{c}{Development rules}\\
\cmidrule(lr){2-2}\cmidrule(lr){3-5}
 & Natural & Natural & Diagnostic item & QAcc\\
\midrule
Qwen Base, native BF16 & 39.58 & 34.90 & 33.07 & 0.00\\
Qwen Instruct & 37.63 & 35.03 & 31.51 & 3.13\\
Decider v11 & 43.10 & 39.32 & 38.80 & 3.13\\
StartLux 2B, FP32 research path & 38.28 & 38.54 & 41.67 & 6.25\\
Controlled Base & 38.15 & 35.68 & 33.33 & 0.00\\
LoRA, mean over fit seeds & 77.99 & 56.51 & 56.86 & 18.75\\
Jev & 60.68 & 55.86 & 49.74 & 10.42\\
\bottomrule
\end{tabular*}
\caption{Higher item accuracy coexists with lower all-four correctness across the tested systems. Scores are macro-averaged percentages across the tested tasks. Each system retains its interface and training history.}
\label{tab:initial-system-overview}
\end{table*}

Across the three optimizer seeds, LoRA has standard deviations of 3.50 percentage points in new-rule natural accuracy and 1.04 in quartet QAcc.

\subsection{Replication by Rule Split and State Seed}
\begin{figure*}[t]
\centering
\includegraphics[width=\textwidth]{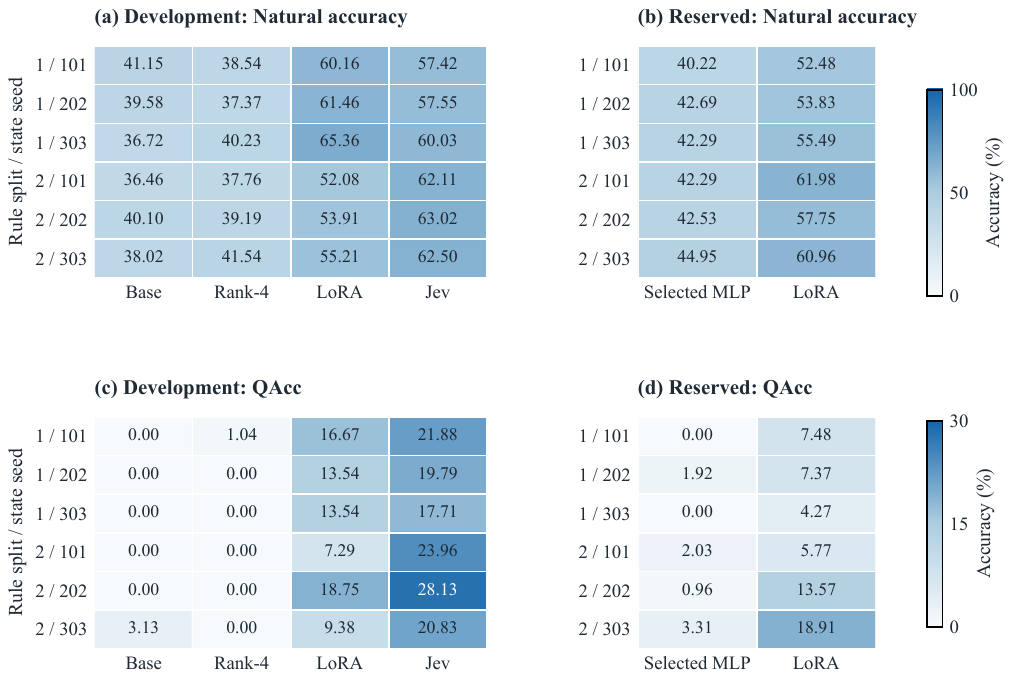}
\caption{LoRA gains recur across rule splits and state seeds, while joint correctness varies across cells. Left panels retain all development systems; right panels compare source-selected MLPs with LoRA on reserved rules. Numbers give accuracy (\%). Natural accuracy uses a shared 0--100 color scale, while QAcc uses a shared 0--30 scale across all development and reserved panels for the corresponding metric, keeping cell comparisons on the same scale.}
\label{fig:replication-cells}
\end{figure*}
\lead{Jev and LoRA change their relative ranking across rule splits.}
Figure~\ref{fig:replication-cells} retains every development cell. LoRA reaches higher natural accuracy on split 1 than on split 2, while Jev shows the opposite ordering. The pooled means summarize these tested allocations; the cell scores show how system rankings change with the assigned source rules.

Tables~\ref{tab:development-cells-full} and~\ref{tab:reserved-models-full} give cell scores from Figure~\ref{fig:replication-cells}.

\begin{table*}[t]
\centering\small
\setlength{\tabcolsep}{4pt}
\begin{tabular}{llrrrrrrrr}
\toprule
\rowcolor{tableblue}
& & \multicolumn{2}{c}{Base} & \multicolumn{2}{c}{Rank-4 readout} & \multicolumn{2}{c}{LoRA} & \multicolumn{2}{c}{Jev}\\
Split & State seed & Natural & QAcc & Natural & QAcc & Natural & QAcc & Natural & QAcc\\
\midrule
1 & 101 & 41.15 & 0.00 & 38.54 & 1.04 & 60.16 & 16.67 & 57.42 & 21.88\\
1 & 202 & 39.58 & 0.00 & 37.37 & 0.00 & 61.46 & 13.54 & 57.55 & 19.79\\
1 & 303 & 36.72 & 0.00 & 40.23 & 0.00 & 65.36 & 13.54 & 60.03 & 17.71\\
2 & 101 & 36.46 & 0.00 & 37.76 & 0.00 & 52.08 & 7.29 & 62.11 & 23.96\\
2 & 202 & 40.10 & 0.00 & 39.19 & 0.00 & 53.91 & 18.75 & 63.02 & 28.13\\
2 & 303 & 38.02 & 3.13 & 41.54 & 0.00 & 55.21 & 9.38 & 62.50 & 20.83\\
\bottomrule
\end{tabular}
\caption{LoRA gains recur across development cells, while the relative ranking of LoRA and Jev changes across rule splits. Scores are percentages, with equal weight assigned to each task family within every evaluated cell.}
\label{tab:development-cells-full}
\end{table*}

\begin{table*}[t]
\centering\small
\setlength{\tabcolsep}{4pt}
\begin{tabular}{llrrrrr}
\toprule
\rowcolor{tableblue}
Split & State seed & \shortstack{Readout\\natural} & \shortstack{Readout\\QAcc} & \shortstack{LoRA\\natural} & \shortstack{LoRA\\QAcc} & Gap (pp)\\
\midrule
1 & 101 & 40.22 & 0.00 & 52.48 & 7.48 & +12.26\\
1 & 202 & 42.69 & 1.92 & 53.83 & 7.37 & +11.14\\
1 & 303 & 42.29 & 0.00 & 55.49 & 4.27 & +13.20\\
2 & 101 & 42.29 & 2.03 & 61.98 & 5.77 & +19.69\\
2 & 202 & 42.53 & 0.96 & 57.75 & 13.57 & +15.22\\
2 & 303 & 44.95 & 3.31 & 60.96 & 18.91 & +16.01\\
\bottomrule
\end{tabular}
\caption{Every fixed LoRA model exceeds its selected readout on reserved-rule natural accuracy. Scores are percentages, and gaps are percentage points with equal weights across all of the reserved evaluation rule blocks.}
\label{tab:reserved-models-full}
\end{table*}

\subsection{Task Profiles}

\begin{figure*}[t]
\centering
\includegraphics[width=\linewidth]{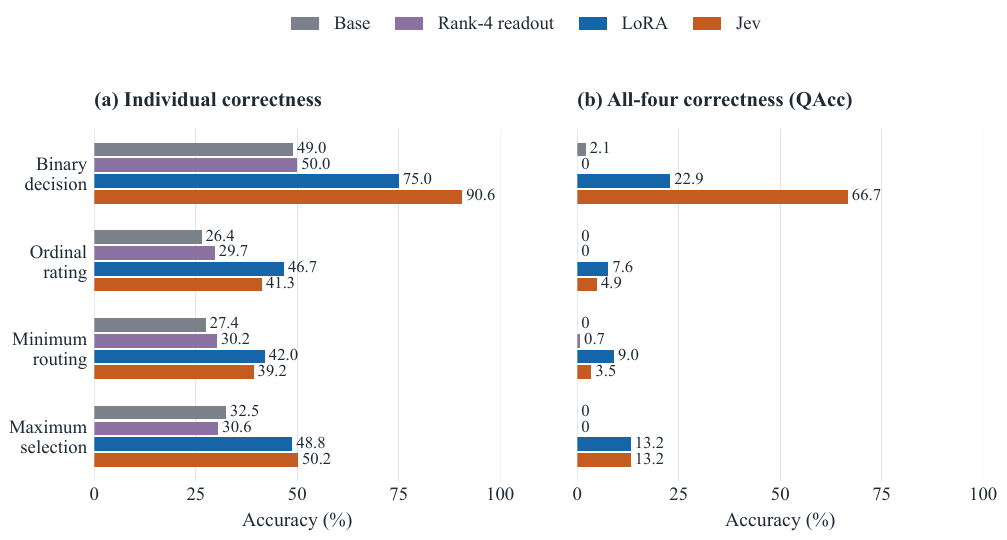}
\caption{Individual correctness and all-four correctness reveal different task profiles. Jev leads in binary QAcc; LoRA leads in rating and routing and ties Jev in selection. Both panels use the same development questions, task/split weighting, and 0--100\% scale; labels give every individual and joint score to one decimal place.}
\label{fig:p4-task-profile}
\end{figure*}
\label{app:p4-tasks}
Figure~\ref{fig:p4-task-profile} reports all task-level diagnostic means. Within each cell, option counts receive equal weight within task, followed by equal state-seed and rule-split weights. Every task/$K$ stratum contains 64 natural and 32 diagnostic questions, including eight complete quartets. Natural items and diagnostic quartets use separately sampled complete states within each model evaluation cell.

\begin{table}[t]
\centering\small
\setlength{\tabcolsep}{4pt}
\begin{tabular}{llrr}
\toprule
\rowcolor{tableblue}
Model & Task & Item accuracy & QAcc\\
\midrule
Base & Binary & 48.96 & 2.08\\
Base & Rating & 26.39 & 0.00\\
Base & Routing & 27.43 & 0.00\\
Base & Selection & 32.47 & 0.00\\
Rank-4 readout & Binary & 50.00 & 0.00\\
Rank-4 readout & Rating & 29.69 & 0.00\\
Rank-4 readout & Routing & 30.21 & 0.69\\
Rank-4 readout & Selection & 30.56 & 0.00\\
LoRA & Binary & 75.00 & 22.92\\
LoRA & Rating & 46.70 & 7.64\\
LoRA & Routing & 42.01 & 9.03\\
LoRA & Selection & 48.78 & 13.19\\
Jev & Binary & 90.62 & 66.67\\
Jev & Rating & 41.32 & 4.86\\
Jev & Routing & 39.24 & 3.47\\
Jev & Selection & 50.17 & 13.19\\
\bottomrule
\end{tabular}
\caption{Task-level item and joint scores distinguish binary strength from performance on ratings and record selection. Percentages give equal weight to option counts, state seeds, and rule splits within each evaluated task.}
\label{tab:task-scores-full}
\end{table}

\subsection{Correct-Count Distributions and Required Response Changes}
\label{app:p1-structure}
\begin{figure*}[t]
\centering
\includegraphics[width=\linewidth]{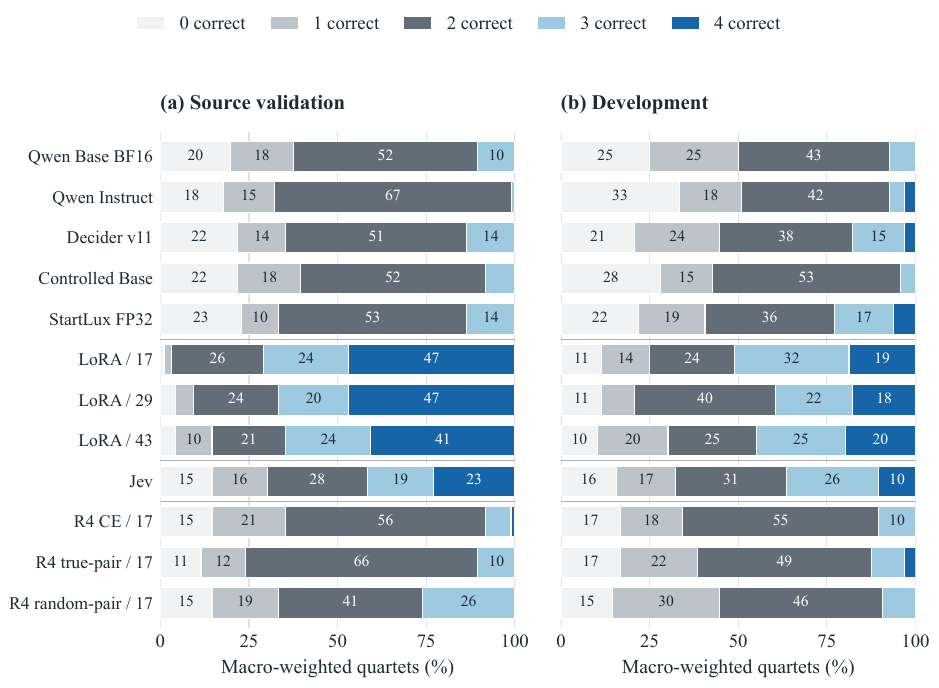}
\caption{Partial correctness occupies much more probability mass than all-four success for most systems. Segment widths retain the macro-weighted percentage of quartets with zero to four correct answers; dark blue marks all-four success. Every model and seed is retained, and labels annotate segments with macro weights of at least 10\%.}
\label{fig:correct-counts}
\end{figure*}

\lead{The quartet analysis records partial success and two distinct failures to respond.}
Figure~\ref{fig:correct-counts} reanalyzes saved initial-batch predictions without new inference. It includes the original model and seed evaluations. Source and development panels each contain 320 questions in 80 complete quartets per system. Rows use the base rendering. Option-reversal scores are not mixed into this analysis. Macro weighting applies to the correct-count distribution; its percentages need not convert directly to unweighted integer counts of evaluated quartets.

\lead{LoRA reduces nonresponse in the initial development sample without eliminating it.}
For seed 17, unchanged semantic answers occur on 35.42\% of required rule switches and 35.94\% of required state switches, compared with Controlled Base's 75.52\% and 84.90\%. LoRA is correct on both members of only 35.94\% of rule-switch pairs and 41.67\% of state-switch pairs. Thus, responsiveness, pair correctness, and QAcc remain different behavioral quantities. The saved reassociation references preserve within-stratum correctness marginals. They are descriptive controls, not causal baselines or an IID statistical sampling model.

\section{Information and Answer-Binding Controls}
\label{app:binding}
\subsection{Controlled Input Manipulations}
D0 preserves the original diagnostic item. D1 adds the supplementary-information instruction without supplying new values. D2 appends a table of every rule's oracle value on every record. D3 uses the identical formula/record-key structure with each numeric entry replaced by \texttt{unavailable}. D4 supplies values for the requested rule only, with D5 as its missing-value counterpart. D6 supplies the correct semantic answer text, with D7 replacing it by \texttt{unavailable}. Conditions D1--D7 remove the instruction to ignore other fields. Their shared instruction specifies the exact formula, thresholds, tie-break, and binding of supplied semantic answers to displayed options. Missing entries instruct the model to use the original input facts or records.

D2 and D3 preserve original coordinates, question, options, and gold answer. The table is ordered by formula and includes every rule in the relevant pool. Questions sharing a state therefore receive the same table across rule changes. In replication this covers eight development rules. D4 and D6 contain question-dependent information and therefore alter the information held fixed across a rule switch. All auxiliary conditions are reported separately from original-task accuracy and were not used in the training of any of the evaluated models.

D4 and D6 reveal question-specific information, so their auxiliary all-four scores do not test the original rule/state response requirement. Table~\ref{tab:value-scores-full} reports the aggregate value effects across all tested conditions. The following positive controls examine the binding to the explicitly supplied answer.

\subsection{Replicated Auxiliary Effects}
\label{app:replicated-auxiliary}
\lead{The all-rule value-table benefit varies strongly across the four measured systems.}
Figure~\ref{fig:values} summarizes D2/D3 in the replication development batch. Each of the six cells contributes 320 diagnostic items paired between D2 and D3 within that cell. Aggregation averages $K$ within task, tasks within cell, the three state seeds within a rule split, and finally the two rule splits. Across cells, LoRA item/QAcc differences range from $-0.26$ to $7.03$ pp and $-3.13$ to $9.38$ pp. Jev ranges are $30.47$--$38.02$ pp and $50.00$--$59.38$ pp. These are two rule allocations with three state samples each. They are not six independently sampled rule replications.

\begin{table*}[t]
\centering\small
\setlength{\tabcolsep}{4pt}
\begin{tabular}{lrrrr}
\toprule
\rowcolor{tableblue}
& \multicolumn{2}{c}{Item accuracy} & \multicolumn{2}{c}{QAcc}\\
Model & Missing values & Complete values & Missing values & Complete values\\
\midrule
Base & 35.72 & 40.19 & 0.00 & 2.08\\
Rank-4 readout & 35.81 & 39.28 & 0.00 & 3.13\\
LoRA & 49.48 & 52.95 & 7.47 & 11.98\\
Jev & 54.43 & 90.63 & 20.31 & 75.69\\
\bottomrule
\end{tabular}
\caption{Complete values improve joint correctness most for Jev on the paired development inputs. Scores are percentages under the same task, option-count, state-seed, and rule-split macro weighting used in Figure~\ref{fig:values}.}
\label{tab:value-scores-full}
\end{table*}

\subsection{Answer-Binding Positive Controls}
Answer-binding positive controls, in percent. All conditions reveal the correct semantic answer. Full/minimal varies context, and legacy/direct varies instruction framing. The 80 source-validation indices retain repeated semantic inputs, so the denominator counts evaluated items rather than distinct, independently drawn semantic states.

\begin{figure}[t]
\centering
\includegraphics[width=\linewidth]{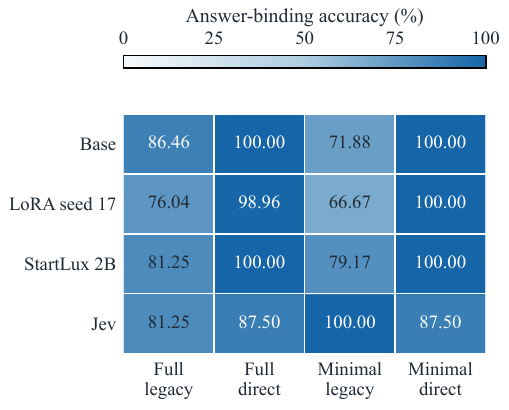}
\caption{Answer binding depends on context and instructions even when the correct semantic answer is supplied. Direct instructions raise local-model accuracy, while Jev performs best with minimal context and legacy instructions. Numbers give accuracy (\%) using a single 0--100 sequential color scale shared across all systems, context lengths, and instruction conditions.}
\label{fig:binding-controls}
\end{figure}
The binding study uses eight item indices per task/option-count layer, selected from source natural validation with recorded answer rotation and identity ordering. The direct instruction removes the original task question and asks the model to bind the supplied answer to an option. Under minimal/direct conditions there are only 31 unique visible inputs per system, with repeated groups tracked separately. Local visible-input identities use effective token IDs. Jev identities use request bodies. These fingerprints are not interchangeable. The repeated groups have consistent semantic choices in this study, although some probability values vary.

\lead{Correct-answer hints and value tables are not a monotone difficulty ladder.}
In the original diagnostic batch, Jev is correct under D2 but wrong under D6 on 21.88\% of macro-weighted items. The reverse occurs on 4.17\%. Providing the right answer does not guarantee that the interface will use it. The binding study is a source-validation positive control. Its source-validation rates and the original new-rule D6 rates use different evaluation inputs, preventing a direct subtraction that isolates the effect of changing instruction wording alone.

\section{Source-Competence and Optimization Audits}
\label{app:source-audit}
The operational source gate requires natural accuracy of at least 90\% and all-four diagnostic accuracy of at least 75\%, on both original source-training questions and independent source-audit questions. These thresholds define the premise needed for the proposed source-competent transfer analysis. They are not universal standards for rule understanding. Selection uses only source-natural validation NLL, with the gate evaluated afterward. The gate is checked separately on original and reversed-option source-training renderings.

\lead{Selected richer readouts fail the source-competence gate.}
All selected-configuration runs fail, including refits with seeds 17, 29, and 43. Across these runs, source-training natural accuracy ranges from 56.64\% to 68.00\%, source-audit natural accuracy from 41.80\% to 52.08\%, and audit QAcc from 0\% to 7.29\%. These are descriptive ranges over a finite set of fits. The six seed-17 models later used in the reserved-rule evaluation are a subset, with source-audit natural accuracy of 41.80--49.87\% and all-four QAcc of 0--3.13\%.

\section{Prospective Evaluation on Reserved Rules}
\label{app:blocks}
\subsection{Frozen Models, Blocks, and Coverage}
After development, the six selected seed-17 richer readouts and six replication LoRA checkpoints are frozen. All 26 reserved semantic components are ordered by SHA256 of a fixed prefix and component ID, then paired adjacently into 13 blocks without rule reuse. For every block and task/$K$ stratum, 16 new natural states are drawn before labels are computed and evaluated under both rules. Thus the natural evaluation has $13\times10\times16\times2=4{,}160$ questions. All rule blocks and models remain fixed.

Two diagnostic quartets per block/stratum are attempted under the fixed eligibility rule. The 254 feasible quartets out of 260 planned contain 1,016 diagnostic items. Six missing quartets come from block 11 binary/$K=2$, block 11 ordinal/$K=3$, and block 12 binary/$K=2$, with two omitted quartets in each stratum. They remain coverage omissions. They do not trigger replacement rules or affect the primary natural evaluation. The combined 5,176 questions use 2,588 new complete states, disjoint from the 43,140-state historical ledger and disjoint across blocks. Both rules in each block are evaluated on the same natural states, retaining paired model comparisons within that rule block.

\begin{figure*}[t]
\centering
\includegraphics[width=\linewidth]{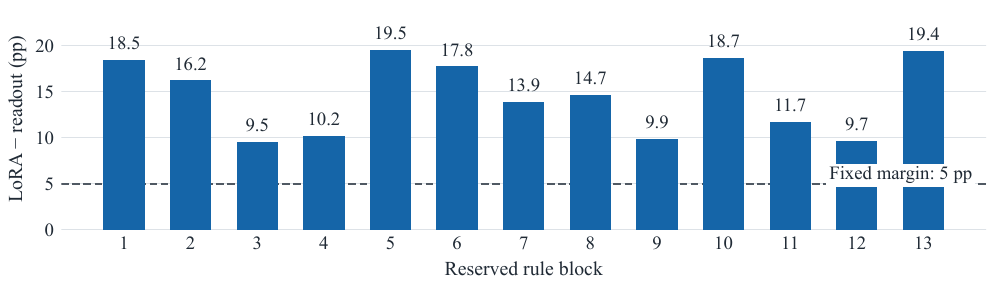}
\caption{LoRA exceeds the prospectively fixed 5 pp margin over the selected readout in every reserved block. Bars begin at zero and labels give the natural-accuracy difference (pp); the dashed line marks the decision margin.}
\label{fig:reserved-block-deltas}
\end{figure*}

\subsection{Finite-Benchmark Decision Rule}
Let $A_{b,m}^{L}$ and $A_{b,m}^{H}$ be macro natural accuracy for LoRA and the frozen readout in model pair $m$ on block $b$. The prospectively specified LoRA--readout contrast is
\begin{equation}
 \Delta_b=\frac{1}{6}\sum_{m=1}^{6}
 (A_{b,m}^{L}-A_{b,m}^{H}),\qquad
 \bar\Delta=\frac{1}{13}\sum_{b=1}^{13}\Delta_b.
\end{equation}
Before generating reserved-rule questions or predictions, the protocol requires $\bar\Delta\ge0.05$ and at least ten of thirteen blocks to satisfy $\Delta_b>0.05$. Base scores, diagnostic QAcc, and task-specific results are secondary descriptions. A failure of either primary condition would be reported as failing the fixed criterion, without extending the experiment.

\lead{The fixed LoRA--readout comparison meets its predeclared finite-benchmark criterion.}
The mean difference is 14.59 percentage points, and all 13 block differences exceed five points, ranging from 9.51 to 19.53 points. Across-block means for the six separate model pairs range from 11.14 to 19.69 points. Averaging the three pairs from each source split gives 12.20 and 16.97 points. Figure~\ref{fig:reserved-block-deltas} retains all thirteen individual rule-block effects.

Blocks do not reuse rules or complete states, but they share trained models and a finite grammar. The six model pairs also share this reserved evaluation. The pass criterion is consequently a statement about the reserved finite benchmark, not a population-level significance test or independent source-retraining confirmation. No IID row bootstrap, binomial block probability, or population $p$-value is assigned to this result. The criterion evaluates transfer by these fixed procedures on the prospective reserved-rule transfer benchmark.

Figure~\ref{fig:replication-cells} reports each of the six fixed model pairs.

\section{Probes and Counterfactual Rule Interventions}
\label{app:mechanisms}
\subsection{Probe Data, Targets, and Fitting}
Base and state-101 LoRA are tested in both splits. Each model/split cell has 2,880 source-training, 640 source-validation, 640 independent source-audit, and 640 development questions. Features are layers 8/16/24 at the final input and final question positions, plus input-embedding segment means. Source-training standardization uses a $10^{-6}$ scale floor. A feature-name-seeded Gaussian projection reduces each feature to 256 dimensions, scaling entries by the inverse square root of input dimension.

Value targets are $\operatorname{asinh}(f_r(x_j,y_j,z_j)/10)$, padded to eight columns and masked by valid records. Separate answer targets are eight-way one-hot vectors. Weighted ridge regression fits each column with an unpenalized intercept, renormalizing source-training weights over valid rows. Source-validation macro masked MSE selects L2 from $\{0.1,1\}$. Error averages valid columns within items, then task/$K$/rule strata. Value probes reconstruct decisions by sign, transformed thresholds, or min/max selection. Answer probes predict the highest-scoring legal candidate among the candidate answers supplied for each evaluated task family and its prescribed legal option count.

\begin{figure*}[t]
\centering
\includegraphics[width=\linewidth]{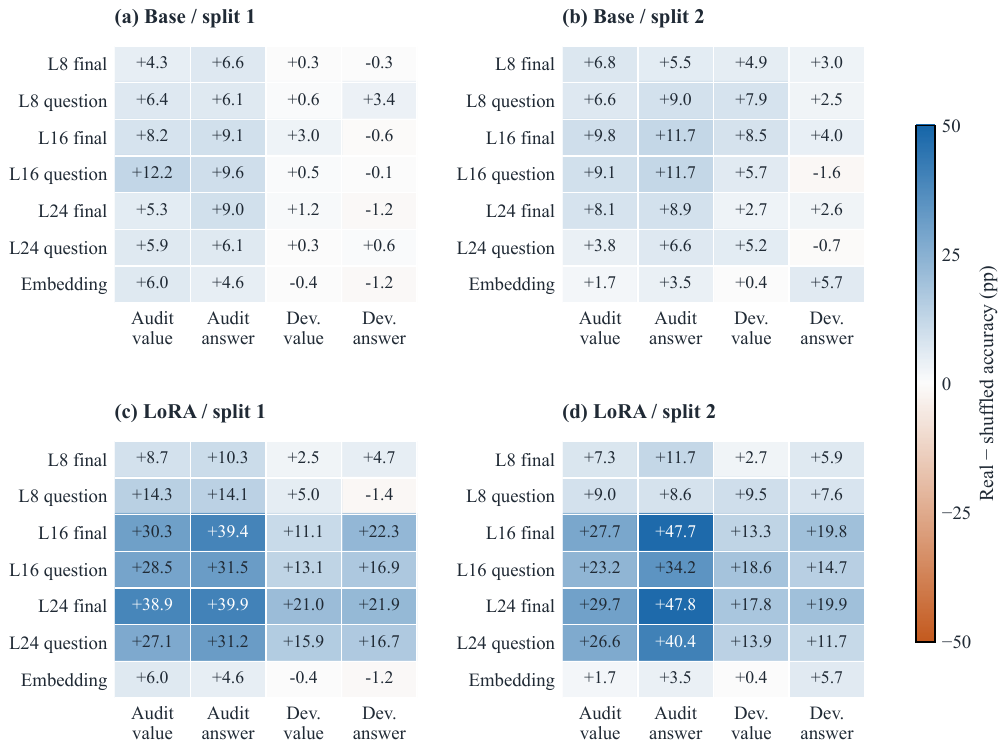}
\caption{Probe gains over shuffled targets are stronger on the source audit than on development inputs at many LoRA locations. Numbers give real minus shuffled-target accuracy (pp) for value and answer targets at every tested location. The diverging scale is centered at zero and shared across every evaluated layer and model panel.}
\label{fig:probes}
\end{figure*}

The shuffled control permutes training labels within task/$K$ with seed 1701 and retains the real-label regularization. Figure~\ref{fig:probes} shows every location. These supervised targets are never inputs to the language model, and recoverable information alone does not establish causal use in answering.

\subsection{Donor--Receiver Pairs and Source Selection}
Each intervention pairs a receiver $(s_r,r_r)$ with a donor $(s_d,r_d)$ from the same task and option-count stratum, requiring both $s_d\ne s_r$ and $r_d\ne r_r$. The counterfactual target is
\begin{equation}
 y^*=g_t(r_d,s_r),
\end{equation}
the donor rule applied to the receiver state. Pair construction uses input identities and the oracle, not a model's predictions. Pairs are stratified by task/$K$ and target-answer equality with donor and receiver answers. Each stratum retains up to four pairs in deterministic order, prioritizing distinct donor, receiver, and counterfactual answers. The fixed source-audit and development evaluations each contain 160 pairs per model/split cell. Binary tasks cannot have three distinct answers and are not treated as isolating this answer property.

Candidate intervention locations are the final question positions at layers 8 and 16. Source-training diagnostic pairs provide same-state/different-rule representation differences. Their singular value decomposition supplies a four-dimensional orthonormal basis $B$ from the leading right singular vectors. This defines a rule-associated direction set, not a proven rule-only representation. For receiver and donor hidden vectors $h_r,h_d$, the two candidate patch families are
\begin{align}
 h'_r&=h_d &&\text{(full)},\\
 h'_r&=h_r+BB^\top(h_d-h_r) &&\text{(subspace)}.
 \label{eq:patch-detail}
\end{align}
The amplitude is fixed at one. Source-validation target success selects the layer and mode by maximizing patch success minus the best of native, matched-norm random-direction, and answer-direction success. Exact ties prefer the earlier layer and then the subspace mode. Evaluation retains this source choice. Audit and development inputs evaluate this source-selected intervention without changing its selected layer, mode, or replacement amplitude.

\subsection{Intervention Controls and Numerical Validation}
The controls include native prediction and restoration, the selected patch, and same-rule or same-answer donors. Additional controls use matched-norm random and answer directions, rule-subspace and random erasure, and reversal of candidate order while applying the same selected intervention.

Same-rule donors change the state while retaining the receiver rule. Same-answer donors use another state and rule whose answer matches the receiver's answer. Answer directions subtract source-training answer prototypes within task/$K$ strata and match the selected patch norm. Random directions are deterministically seeded per pair and layer. Rule-subspace erasure subtracts $BB^\top(h_r-\mu)$ using the source-training mean $\mu$. Random erasure uses a perturbation of the same norm. Donors and receivers share a checkpoint and use vectors from the same source-selected layer and token position.

Receivers are captured in the native forward batch, and restoration must retain identical choices and probabilities within $10^{-5}$. Reversed candidates use receivers captured under their own ordering. Source validation selects among layers 8 and 16 and full or subspace replacement, with amplitude fixed at one. Base/split 1 uses full replacement at layer 8; Base/split 2 uses the layer-16 subspace. Both LoRA cells use the layer-8 subspace. These fixed evaluation settings remain source-selected.

\begin{figure*}[t]
\centering
\includegraphics[width=\textwidth]{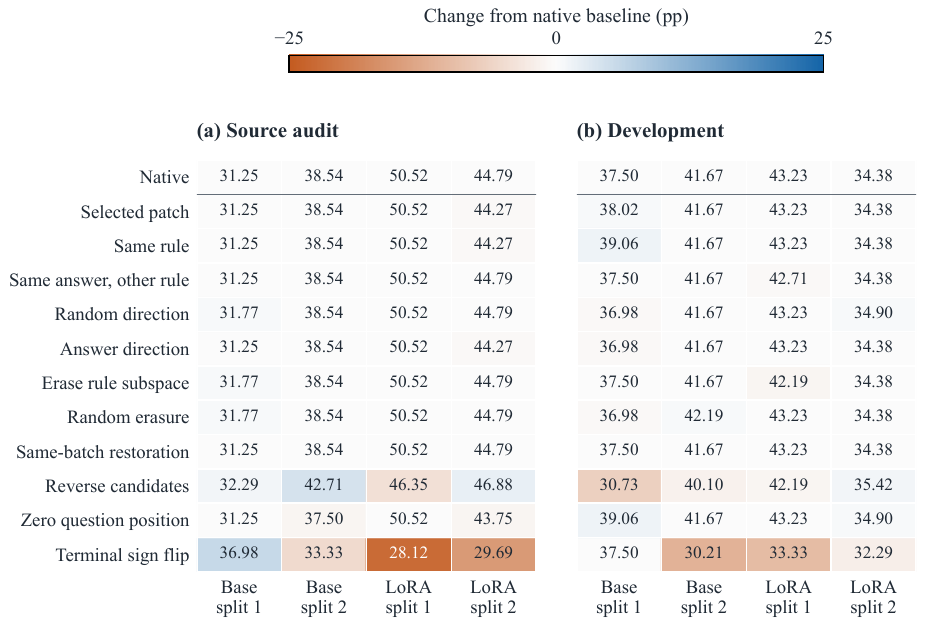}
\caption{Source-selected patches produce little change in counterfactual target success relative to the native baseline. Numbers retain absolute success (\%); shading shows the within-column change from the native row (pp) on a shared, zero-centered scale. Audit and development retain all of the positive and negative controls.}
\label{fig:intervention-controls}
\end{figure*}

Applied BF16 norms, conversion errors, probabilities, and margins characterize the actual perturbation. Zeroing the question position and flipping the terminal sign provide positive controls for path sensitivity. Figure~\ref{fig:intervention-controls} compares counterfactual-target success for the selected patch and the retained donor, direction, erasure, and restoration controls on paired audit and development inputs.

For Base/split 1, Base/split 2, LoRA/split 1, and LoRA/split 2, applied BF16 relative norms are 4.84\%, 3.39\%, 2.19\%, and 4.07\%, respectively. Prediction changes are 1.56\%, 0.00\%, 0.00\%, and 0.00\%. Native and patched target success for both source-rule splits appear separately in Figure~\ref{fig:intervention-controls}.

Each audit/development cell has 160 donor--receiver pairs, macro-averaged over task and option count. The relative norm is $100\|h_r^{\rm applied}-h_r\|_2/\max(\|h_r\|_2,10^{-12})$, averaged with the same task/$K$ weights. The applied vector includes the actual BF16 cast. A value of 4.84 thus denotes a change of 4.84\% relative to the receiver norm.

\section{Joint-Metric Calibration and Conditional Uncertainty}
\label{app:calibration}
\lead{Stratified reassociation characterizes the joint reference implied by the saved prediction marginals.}
Calibration reuses predictions from the fixed models. Within each cell, reassociation preserves prediction marginals conditional on task/$K$, quartet slot, and true label. It uses 1,000 reassociations to obtain descriptive quantile ranges. This reference breaks associations within the original quartets. It is neither an independent-row sampling model nor a causal test of rule execution. The development replication contains 480 complete quartets, and the reserved rules yield 254 feasible quartets. Calibration covers the feasible task/$K$ strata recorded in the fixed diagnostic evaluation sets.

The deterministic controls use only original-rendering source-training natural data. Let $R_S$ be its distinct rules and $S_{t,K}$ its distinct complete states for task $t$ and option count $K$. Let $M$ return the modal answer index, breaking ties toward the smallest index. The source-label-majority control returns the mode of source labels within $(t,K)$. The other controls are
\begin{align}
 h_{\rm rule\mbox{-}blind}(s,t,K)&=M\{g_t(r,s):r\in R_S\},\\
 h_{\rm state\mbox{-}blind}(r,t,K)&=M\{g_t(r,s):s\in S_{t,K}\}.
\end{align}
Oracle answers use the original option-index convention, including record-ID tie-breaking. Thus the rule-blind control votes over source rules on the presented state, while the state-blind control votes over source states under the presented rule. Neither uses development labels to fit or select predictions.

On these diagnostic questions, source-label-majority, rule-blind, and state-blind controls obtain item accuracies of 34.20\%, 40.32\%, and 33.64\%, with zero QAcc. Uniform independent choices give 32.72\% item accuracy and 2.79\% QAcc. Paired intervals use 2,000 bootstrap resamples with fixed seeds. Natural questions that share a complete state remain grouped, and diagnostic questions are resampled as complete quartets. Models and conditions share resampling weights, retaining the original task/$K$ and cell strata. Reserved-rule comparisons first compute metrics within rule blocks and then average equally over fixed models and blocks. These intervals are conditional on the existing rule splits, models, and evaluation samples. They do not cover new rule allocations, training seeds, or API service variation, and retain the dependence among cells that share the same source-rule allocation.

\lead{Maximum-selection quartets show positive departures from the marginal reference.}
Table~\ref{tab:calibration-task} separates tasks while preserving the original stratification. Both LoRA and Jev exceed the maximum-selection reference range. Other task-level observations lie within their ranges. The pooled result therefore cannot be interpreted as absence of joint dependence in every task. Small, singleton, and constant-prediction strata limit how much the reference can change. Appendix~\ref{app:calibration-support} reports their counts and metric weights, alongside maximum-selection scores for each evaluated cell and option count.

Task-specific development calibration, averaging six fixed cells. References independently reassociate slots within cell/task/$K$/gold strata (1,000 draws, seed 2026100802). The ranges summarize the reassociated reference distribution. They are descriptive quantiles, rather than confidence intervals or multiplicity-adjusted significance assessments.

\begin{table*}[!htbp]
\centering\small
\setlength{\tabcolsep}{3pt}
\begin{tabular}{llcccc}
\toprule
\rowcolor{tableblue}
Task & Model & QAcc & Ref. mean & Ref. 95\% range & Obs. minus ref.\\
\midrule
binary decision & Base & 2.08 & 4.29 & [0.00, 10.42] & -2.20\\
binary decision & Jev & 66.67 & 66.55 & [62.50, 70.83] & 0.11\\
binary decision & LoRA & 22.92 & 27.90 & [20.83, 35.42] & -4.98\\
minimum routing & Base & 0.00 & 0.07 & [0.00, 0.69] & -0.07\\
minimum routing & Jev & 3.47 & 3.00 & [0.69, 5.56] & 0.47\\
minimum routing & LoRA & 9.03 & 5.90 & [3.45, 9.03] & 3.13\\
ordinal rating & Base & 0.00 & 0.00 & [0.00, 0.00] & 0.00\\
ordinal rating & Jev & 4.86 & 6.80 & [4.17, 9.72] & -1.93\\
ordinal rating & LoRA & 7.64 & 7.34 & [4.86, 9.74] & 0.30\\
maximum selection & Base & 0.00 & 0.00 & [0.00, 0.00] & 0.00\\
maximum selection & Jev & 13.19 & 8.75 & [6.25, 11.81] & 4.45\\
maximum selection & LoRA & 13.19 & 7.60 & [4.17, 10.42] & 5.60\\
\midrule
\multicolumn{6}{l}{Pooled, ordered rule pairs}\\
All tasks & LoRA & 13.19 & 13.33 & [12.15, 14.58] & -0.13\\
All tasks & Jev & 22.05 & 21.69 & [20.66, 22.92] & 0.35\\
\multicolumn{6}{l}{Pooled, rule pairs and state vectors}\\
All tasks & LoRA & 13.19 & 12.67 & [12.15, 13.19] & 0.52\\
All tasks & Jev & 22.05 & 21.61 & [20.66, 22.57] & 0.44\\
\bottomrule
\end{tabular}
\caption{Joint-score departures depend on task and on the associations retained by reassociation. Task-specific QAcc is compared with 1,000-draw reference ranges; pooled rule-pair and state-vector controls use 2,000 draws. Scores are percentages. The reported ranges describe the reference distribution for each set of preserved associations.}
\label{tab:calibration-task}
\end{table*}

\subsection{References Preserving Rule Pairs and State Responses}
\label{app:pairing-controls}
We reuse the 480 development quartets and compare two additional references with the primary slot-and-label reference. The first also conditions on the ordered pair of rule identities. The second shuffles the two predictions on a state together within the same rule-pair and ordered-label strata, independently for the two state positions. This preserves the observed response association within each state while breaking its link to the other state. It does not match latent state difficulty. Each reference uses 2,000 draws with independently derived seeds for each cell, task/$K$, and reference. Aggregation keeps equal weights over the fixed cells, option counts within tasks, and the four tasks.

\lead{Preserving rule pairs changes the joint-correctness reference.}
The pooled controls appear in Table~\ref{tab:calibration-task}. In maximum selection, retaining rule pairs brings the reference means closer to observed QAcc for both LoRA and Jev. Some finer strata have no movable predictions or a constant QAcc, so adding conditions can also reduce the variation available to the reference. These comparisons characterize dependence on the existing pairing. The evaluation retains the constructed state pairs; it does not test quartets formed from new and independently sampled second complete states.

\section{Common Source Audit and Bounded Competence Matching}
\label{app:source-match-review}
The common audit uses 640 source-natural questions and 320 diagnostic items, forming 80 quartets per cell. These inputs are independent of training and model selection and shared by all methods. The prespecified source-competence gate requires natural accuracy of 90\% and QAcc of 75\% separately on the original training inputs and the independent, held-out source-rule audit evaluation questions.

On the common source audit, Base reaches natural accuracy / QAcc of 36.31\% / 0.00\%. The selected readout reaches 45.70\% / 1.39\%, and LoRA reaches 75.43\% / 35.94\%. None of the three methods satisfies the prespecified competence gate in any of the six fixed replication evaluation cells.

\begin{figure*}[t]
\centering
\includegraphics[width=\textwidth]{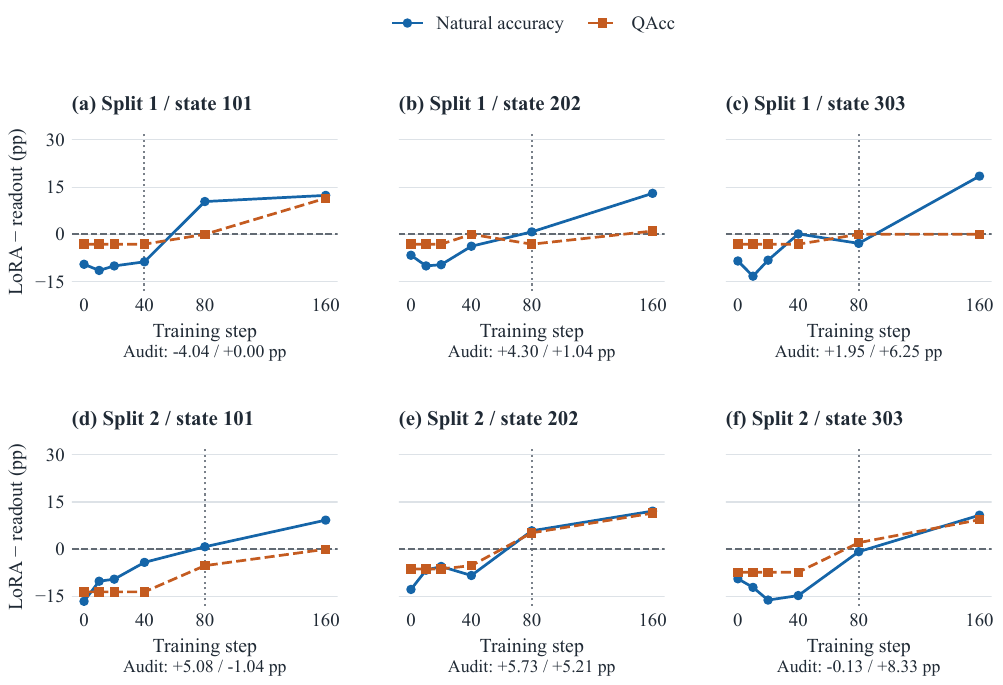}
\caption{Validation trajectories approach the frozen readout at selected early steps, but independent audit gaps remain. Curves show saved LoRA--readout gaps (pp); dotted vertical lines mark selected steps and the horizontal zero line denotes equality. Below-panel labels give audit natural-accuracy / QAcc gaps at those selected steps.}
\label{fig:source-match-curves}
\end{figure*}

Bounded matching uses the original LoRA recipe and training seed 17 for 160 steps per cell. Selection uses the prespecified checkpoints at steps 0/10/20/40/80/160. Selection requires absolute differences from the frozen readout of at most 3 percentage points on both source-validation natural Acc and QAcc. Otherwise, the closest checkpoint is reported and the search stops. The source-validation condition holds in 2/6 cells, but independent source-audit matching succeeds in 0/6. Neither audit nor development data are used for selection or to relax the tolerance. Figure~\ref{fig:source-match-curves} shows all six recorded checkpoints and the independent source-audit gaps for the individual checkpoints selected by source-validation score matching.

Development performance of all closest checkpoints is only a descriptive diagnostic. Its average difference cannot be interpreted as a transfer effect after controlling source competence, nor can unmatched cells be included in a matched group to support such a conclusion. The reproduction materials retain complete checkpoint trajectories and intervals conditional on the source-audit and development inputs used to compare these models.

\section{Equivalent Rewrites, Value Mappings, and Paired Answer Transitions}
\label{app:review-controls}
\subsection{Fixed Inputs and Semantic Verification}
The supplement uses state seed 101 from both development splits. Complete quartets are selected by a fixed SHA256 ordering within task/$K$/rule-pair strata: 40 quartets and 160 questions per split, totaling 80 quartets and 320 questions. Selection does not inspect predictions. Base, the corresponding LoRA, and Jev share the questions. The functional rewrite uses named operations such as \texttt{sum} and \texttt{difference}. The verbal rewrite replaces arithmetic symbols with words inside fully parenthesized expressions. For the archived rule $min(x,y)-(y+z)$, the functional expression is \texttt{difference(minimum(x, y), sum(y, z))}, and the verbal expression is \texttt{(min(x, y) minus (y plus z))}. Both questions ask whether this value is strictly positive and parse back to the original AST. States, thresholds, tie-breaking rules, displayed options, and original gold answers are unchanged.

The correct value table retains the original D2 input. Formula-row mismatch misassigns complete computed-score rows to formulas, while record-column mismatch applies one common column permutation to every formula. Both operations preserve the table's value multiset and serialized character count. A nonidentity column permutation is impossible for single-record tasks, so their 128 questions are marked not applicable. Of the remaining 192 questions, 190 have table targets that conflict with the original oracle. Formula-row mismatch produces 220 conflicting questions.

The target-value counterfactual changes one table value for the requested formula so that the table target moves to the next option in a predefined cycle. Construction uses only the oracle, not model predictions. All 320 questions change their table targets. Value-range, string-length, and sign properties are matched in 292, 165, and 162 questions, respectively, and the remaining differences are retained. This table depends on the specific question, so shared-state QAcc is not reported for this condition. The original task gold answer is always retained, and following the incorrect table target does not count as correct on that original task.

\subsection{Paired Answer Transitions and Denominators}
Define the conflict set $C$ using only questions for which the intervention applies and the original oracle answer $o_i$ differs from the manipulated table target $t_i$. Let $p_i^0$ and $p_i^1$ denote predictions with the correct and incorrect tables. Set $N=|C|$, $B=\sum_{i\in C}\ind[p_i^0=o_i]$, and
\begin{equation}
 T=\sum_{i\in C}\ind[p_i^0=o_i,\ p_i^1=t_i].
\end{equation}
We report the paired-transition proportion over all conflicting questions, $T/N$, and the conditional proportion among initially correct questions, $T/B$, with integer numerators and denominators. Strata with $B=0$ are undefined. The main-text table uses pooled micro estimates. The supplementary macro estimate first pools counts from both splits within task/$K$, then averages equally over $K$ within tasks and over tasks. Per-cell strata and complete $3\times3$ transition matrices are saved separately. Pooled micro, this macro estimate, and equal-cell estimates are distinct. Matrix categories are the original answer, the new table target, and other answers, separating answers that remain correct, follow the table target, or select another incorrect answer option.

Itemwise paired answer transitions for all models. $T/N$ requires the two predictions jointly to satisfy the transition event, while $T/B$ conditions on conflicting questions answered correctly with the correct table. Integer denominators are reported alongside pooled micro-average transition rates.

\begin{table*}[t]
\centering\small
\setlength{\tabcolsep}{4pt}
\begin{tabular}{llccccc}
\toprule
\rowcolor{tableblue}
Model & Intervention & $N$ & $B$ & $T$ & $T/N$ (\%) & $T/B$ (\%)\\
\midrule
Base & Formula-row mismatch & 220 & 69 & 8 & 3.64 & 11.59\\
Base & Record-column mismatch & 190 & 66 & 5 & 2.63 & 7.58\\
Base & Target-value counterfactual & 320 & 116 & 12 & 3.75 & 10.34\\
LoRA & Formula-row mismatch & 220 & 90 & 17 & 7.73 & 18.89\\
LoRA & Record-column mismatch & 190 & 84 & 21 & 11.05 & 25.00\\
LoRA & Target-value counterfactual & 320 & 147 & 19 & 5.94 & 12.93\\
Jev & Formula-row mismatch & 220 & 191 & 169 & 76.82 & 88.48\\
Jev & Record-column mismatch & 190 & 177 & 173 & 91.05 & 97.74\\
Jev & Target-value counterfactual & 320 & 282 & 243 & 75.94 & 86.17\\
\bottomrule
\end{tabular}
\caption{Directed answer transitions distinguish following a supplied target from an undirected answer change. Counts preserve the applicable conflict set: $T/N$ uses all conflicts and $T/B$ conditions on original correctness.}
\label{tab:answer-transfer-full}
\end{table*}

\section{Execution and Reproducibility}
\label{app:reproducibility}
The runtime uses PyTorch 2.8.0+cu128, Transformers 5.18.0, and PEFT 0.21.2 on RTX 4090 GPUs. Local inference follows the precision and padding settings in Appendices~\ref{app:precision} and~\ref{app:numerical-controls}. Protocols, generators, executable oracles, model identities, selected checkpoints, and itemwise predictions have SHA256 identities. BF16 training replay is not bitwise identical. Reloaded readouts retain semantic predictions, with a maximum checked probability difference of $8.35\times10^{-7}$. Jev responses identify \texttt{jev-1.13.0}; successful replacements count once per logical request. The reproduction package excludes model weights and private credentials.

\section{Additional Behavioral Diagnostics}
\label{app:round2}
\subsection{Data Roles and Fixed Evaluation}
These posthoc diagnostics reuse the inspected reserved benchmark without fitting or selection: 4,160 natural questions and 1,016 diagnostic items (254 quartets) in 13 blocks. Two AST-equivalent rewrites yield 72,464 predictions for Base and six fixed LoRA instances. The latter represent two source-rule splits crossed with three state seeds.

All wordings use batch 8 with fixed weights, order, scoring code, BF16 body, and FP32 candidate projection. The 36,232 original-text predictions align with 72,464 hash-verified rewrite predictions. Changing original-text batch size from 16 to 8 changes 101/5,176 Base choices (1.95\%) and 27--42/5,176 choices per LoRA (0.52--0.81\%). Mean absolute logit differences are 0.0322--0.0408, with maximum 0.7095. Flipped items have smaller original top-two margins. Using a baseline at the same batch size shifts both pooled gap changes by $-0.28$ pp. The verbal-rewrite contrast moves from $+0.19$ to $-0.09$ pp, while preserving all eight task-level directions of the measured wording contrasts.

Across models and rewrites, 96.06--98.92\% of wording flips lie outside the observed original-text batch-flip set. We decompose each LoRA--Base gap change by whether either model flips its original-text choice across batches, retaining full-set block/task/$K$ weights and denominators. Batch-stable contributions preserve all eight taskwise directions. Stable/sensitive contributions are $-2.27/-0.37$ pp for the functional rewrite and $+0.22/-0.31$ pp for the paired verbal rewrite.

Intervals use 2,000 resamples (seed 2026100801) within block/task/$K$, preserving paired-rule shared-state clusters for natural questions and complete diagnostic quartets. Sampled indices are shared across conditions and models. Aggregation equally weights $K$ within task, tasks within blocks, blocks, and LoRA instances. These intervals quantify paired input variation while conditioning on the evaluated rules and the fixed, source-selected model checkpoints.

{
\small
\setlength{\tabcolsep}{3pt}
\begin{table*}[t]
\centering
\begin{tabular}{llccccc}
\toprule
\rowcolor{tableblue}
Metric & Models & Original & Rewrite 1 & Rewrite 2 & Change 1 [95\% CI] & Change 2 [95\% CI]\\
\midrule
Natural & All six & 19.30 & 16.66 & 19.21 & -2.64 [-4.80, -0.62] & -0.09 [-1.76, 1.62]\\
Natural & Split 1 & 16.10 & 15.15 & 16.87 & -0.95 [-3.10, 1.10] & 0.77 [-0.87, 2.51]\\
Natural & Split 2 & 22.50 & 18.18 & 21.55 & -4.33 [-6.64, -2.20] & -0.95 [-2.64, 0.80]\\
QAcc & All six & 9.99 & 5.84 & 9.44 & -4.15 [-5.65, -2.80] & -0.55 [-1.44, 0.30]\\
QAcc & Split 1 & 6.37 & 4.88 & 6.16 & -1.50 [-3.13, 0.11] & -0.21 [-1.28, 0.89]\\
QAcc & Split 2 & 13.60 & 6.80 & 12.71 & -6.80 [-9.44, -4.31] & -0.89 [-2.17, 0.39]\\
\bottomrule
\end{tabular}
\caption{Functional rewrites reduce the pooled adaptation advantage; verbal rewrites produce little pooled natural-accuracy change. All entries compare LoRA with Base at batch 8. Changes are paired differences (pp); 95\% intervals condition on the fixed rule allocations and model checkpoints under shared, paired input resampling.}
\label{tab:heldout-rewrites}
\end{table*}
}

{
\small
\setlength{\tabcolsep}{3pt}
\begin{figure*}[t]
\centering
\includegraphics[width=\textwidth]{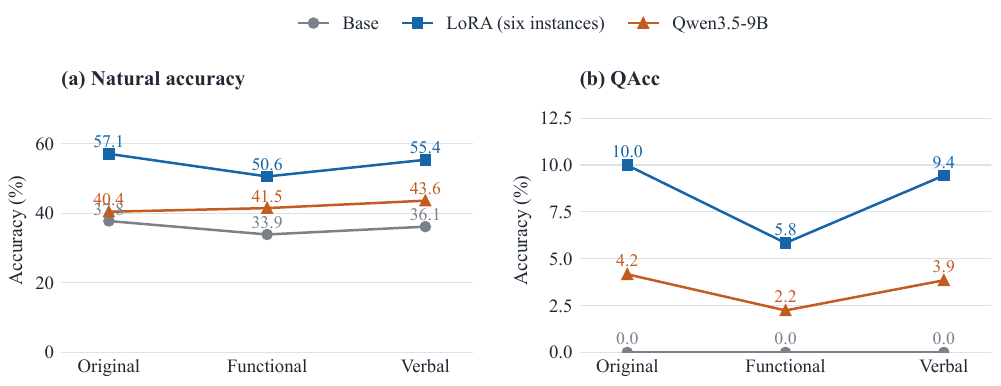}
\caption{LoRA remains above Base across equivalent wordings, with a larger reduction under functional notation. Points connect evaluations of fixed systems at batch 8; labels give accuracy (\%). Natural accuracy and QAcc use separately labeled zero-based scales. LoRA averages six instances, and 9B is a distinct instruction-tuned system.}
\label{fig:wording-systems}
\end{figure*}
}

{
\small
\setlength{\tabcolsep}{3pt}
\begin{table*}[t]
\centering
\begin{tabular}{llccccc}
\toprule
\rowcolor{tableblue}
Task & Model & Original & Rewrite 1 & Rewrite 2 & Change 1 [95\% CI] & Change 2 [95\% CI]\\
\midrule
Binary & Base & 69.95 & 50.96 & 58.17 & -18.99 [-24.77, -12.74] & -11.78 [-17.07, -6.25]\\
Binary & LoRA & 77.72 & 73.60 & 78.04 & -4.13 [-5.77, -2.48] & 0.32 [-0.48, 1.16]\\
Routing & Base & 27.64 & 27.48 & 29.25 & -0.16 [-3.77, 3.37] & 1.60 [-0.88, 4.09]\\
Routing & LoRA & 44.87 & 38.29 & 41.57 & -6.58 [-7.71, -5.46] & -3.30 [-4.22, -2.34]\\
Rating & Base & 22.84 & 23.48 & 24.20 & 0.64 [-0.88, 2.09] & 1.36 [-0.24, 2.96]\\
Rating & LoRA & 54.91 & 48.33 & 53.15 & -6.58 [-7.76, -5.46] & -1.76 [-2.60, -0.95]\\
Selection & Base & 30.61 & 33.65 & 32.93 & 3.04 [-0.32, 6.49] & 2.32 [0.08, 4.57]\\
Selection & LoRA & 50.73 & 42.01 & 48.62 & -8.72 [-9.96, -7.44] & -2.11 [-2.96, -1.27]\\
\bottomrule
\end{tabular}
\caption{Binary-task increases offset adaptation-gain reductions in rating, routing, and selection. Natural accuracy is reported in percentages and paired changes in pp, at batch 8 with complete-state resampling intervals.}
\label{tab:rewrite-tasks}
\end{table*}
}

\subsection{Precision and Padding Controls}
\label{app:numerical-controls}
We select a fixed subset from the existing reserved-rule evaluation using SHA256 ordering with seed 20261009, without consulting predictions. Each task/$K$ stratum contributes eight complete natural state pairs and four diagnostic quartets, totaling 160 natural and 160 diagnostic questions. The original and both rewrites retain identical states, rules, options, and labels. Base and the fixed state-101 LoRA checkpoint from each source split are evaluated under BF16 or FP32 model inference, crossed with dynamic or fixed left padding. Batch size remains eight and item order is unchanged. Fixed padding uses the maximum length across all selected texts, 361 tokens, without truncation. FP32 weights are loaded directly from the original checkpoints, all floating-point model parameters are checked as FP32, and TF32 is disabled. Candidate projection remains FP32 in every condition. The fixed paired design yields 11,520 predictions.

\lead{Wording sensitivity persists when precision and padding are controlled.}
Table~\ref{tab:numerical-controls} separates same-text padding effects from cross-wording disagreement. FP32 substantially reduces the former, while both equivalent rewrites continue to change answers at fixed padding. One LoRA prediction still differs across FP32 padding conditions. The result supports wording sensitivity beyond the observed BF16 padding effect within this paired subset.

Wording rates use the natural questions and average equally over option counts within tasks and then over tasks. Their intervals use 2,000 complete-state-pair bootstrap draws within task/$K$, shared across models and conditions. The diagnostic analysis preserves complete quartets. Unlike the full reserved-rule evaluation, this subset analysis weights the sampled state groups directly within task/$K$, rather than retaining the original rule-block weights. Its intervals condition on the sampled groups and fixed checkpoints. Padding-change counts pool all three wordings and both data types, and therefore use a broader item denominator than the paired natural-question wording comparison.

\begin{table*}[t]
\centering\small
\setlength{\tabcolsep}{4pt}
\begin{tabular}{lcccc}
\toprule
\rowcolor{tableblue}
& \multicolumn{2}{c}{Same-text padding changes} & \multicolumn{2}{c}{Wording disagreement at fixed padding, FP32}\\
\cmidrule(lr){2-3}\cmidrule(lr){4-5}
Model & BF16 & FP32 & Functional (\%, 95\% CI) & Verbal (\%, 95\% CI)\\
\midrule
Base & 26/960 & 0/960 & 44.27 [34.38, 54.69] & 31.25 [23.44, 39.06]\\
LoRA, split 1 & 11/960 & 0/960 & 23.44 [16.67, 30.21] & 12.50 [8.33, 16.67]\\
LoRA, split 2 & 8/960 & 1/960 & 25.00 [19.27, 31.25] & 17.71 [12.50, 22.92]\\
\bottomrule
\end{tabular}
\caption{Wording sensitivity persists with FP32 inference and fixed padding on the controlled subset. Wording-flip rates compare each rewrite with the original on natural questions; same-text padding comparisons retain the wording and measure changes in the selected answer under a different padding condition on identical question inputs.}
\label{tab:numerical-controls}
\end{table*}

\subsection{Nearest-Source Similarity and Conflict Subsets}
\label{app:native-conflict}
For each target rule and task/$K$, the nearest source rule maximizes oracle-answer agreement over uniform finite-domain inputs. Ties within $10^{-14}$ choose the lexicographically smallest source ID. The permutation sensitivity additionally searches the six field permutations, breaking remaining ties lexicographically. This oracle-assisted reference is evaluated on accepted natural states with the target oracle retained as gold. The conflict subset contains questions where their answers differ.

For the two multi-record tasks, records are drawn independently with replacement from the finite domain. Let $(A,B)$ denote the paired rule values of a candidate, with signs reversed for minimum routing so larger is always preferred. Define $p_<(a,b)=P(A<a,B<b)$ and $p_\le(a,b)=P(A\le a,B\le b)$. Since smaller record IDs win ties, exact winner agreement for $K$ records is
\begin{equation}
\begin{aligned}
P(\text{same winner})&=\mathbb E_{(A,B)}[S_K(A,B)],\\
S_K(a,b)&=\sum_{i=0}^{K-1}p_<(a,b)^i p_\le(a,b)^{K-1-i}.
\end{aligned}
\end{equation}
The empirical joint distribution enumerates all finite-domain records, so the formula integrates all ordered $K$-record draws without enumerating their Cartesian power. Records before a candidate must be strictly worse under both rules. Later records may tie the candidate under either rule because record-ID tie-breaking favors the earlier candidate.

The subset-normalized conflict score averages within each observed task/$K$ subset, then equally over $K$ and tasks. It does not equalize rules after filtering. For the additive decomposition, let $g=(\text{rule},t,K)$, $w_g$ be its full-set macro weight, $p_{C,g}$ its conflict fraction, and $\Delta_{C,g},\Delta_{\bar C,g}$ its conditional LoRA-minus-Base differences. Then
\begin{equation}
 \begin{aligned}
\Delta_{\rm full}&=\Delta_C+\Delta_{\bar C},\\
\Delta_C&=\sum_g w_gp_{C,g}\Delta_{C,g},\\
\Delta_{\bar C}&=\sum_g w_g(1-p_{C,g})\Delta_{\bar C,g}.
\end{aligned}
\end{equation}
Weights are $w_g=(4\,|\mathcal K_t|\,26)^{-1}$ for this balanced natural set. Empty subsets contribute zero mass. Their conditional scores are undefined rather than imputed. The two contributions exactly recover the original batch-16 macro advantage. A full-weight conditional conflict score divides its contribution by $\sum_gw_gp_{C,g}$ and differs from the subset-normalized macro computed after conflict filtering (Table~\ref{tab:nearest-weight-details}).

{
\small
\setlength{\tabcolsep}{3pt}
\begin{table*}[t]
\centering
\begin{tabular}{llcccccc}
\toprule
\rowcolor{tableblue}
Mapping & Split & Conflict $n$ & Mass (\%) & \shortstack{Conditional\\weighted} & \shortstack{Conditional\\subset} & \shortstack{Contribution\\conflict} & \shortstack{Contribution\\other}\\
\midrule
native & 1 & 1907 & 43.73 & -2.43 & -3.11 & -1.06 & 16.93\\
native & 2 & 1722 & 38.90 & 12.62 & 11.37 & 4.91 & 17.26\\
permuted & 1 & 1595 & 35.96 & 1.91 & 1.70 & 0.69 & 15.18\\
permuted & 2 & 1454 & 32.97 & 14.74 & 13.70 & 4.86 & 17.31\\
\bottomrule
\end{tabular}
\caption{Nearest-source nonconflicts account for most of the full-set adaptation gain. Conditional gains and weighted contributions are in pp. Weighted estimates retain full-set conflict mass; subset estimates recompute the task/$K$ means on retained questions before averaging scores equally across strata within each task family.}
\label{tab:nearest-weight-details}
\end{table*}
}

{
\small
\setlength{\tabcolsep}{3pt}
\begin{table*}[t]
\centering
\begin{tabular}{llcc}
\toprule
\rowcolor{tableblue}
Mapping & Split & Shared conflict $n$ & LoRA--Base (pp)\\
\midrule
native & 1 & 918 & -0.77\\
native & 2 & 918 & 10.08\\
permuted & 1 & 727 & 7.11\\
permuted & 2 & 727 & 10.12\\
\bottomrule
\end{tabular}
\caption{The common conflict intersection permits both source splits to be compared on identical questions. All ten task/$K$ strata remain represented; percentages use the subset-normalized task and option-count macro.}
\label{tab:common-conflict}
\end{table*}
}

Per-rule agreement records are retained with the offline analysis. Uniform-domain agreement and accepted-sample agreement remain separate: exclusion of historical states and finite sampling can change the latter. Selection and model-performance correlations are exploratory. None is used to redefine the source or target partitions.

\setlength{\textfloatsep}{12pt plus 2pt minus 2pt}
\setlength{\dbltextfloatsep}{12pt plus 2pt minus 2pt}
\setlength{\floatsep}{8pt plus 2pt minus 2pt}
\setlength{\dblfloatsep}{8pt plus 2pt minus 2pt}
\subsection{Agreement-Stratified Transfer and Complete-Quartet Conflict Loads}
\label{app:agreement-strata}
\lead{Lower oracle agreement separates weaker transfer from the pooled adaptation gain.}
This posthoc analysis reuses all saved batch-16 reserved predictions: 4,160 natural questions and 254 complete quartets. It adds no training, inference, or service requests. For each target pair, we average its two native nearest-source agreements over the uniform finite domain. Within each source split and task/$K$, the lower, middle, and upper thirds define low, middle, and high agreement. Ties remain together. These are relative groups; their absolute agreement ranges can overlap across tasks. Each task/$K$ allocates 5, 4, and 4 of its 13 target pairs to these groups. All ten task/$K$ strata remain represented in each pooled oracle-agreement group.

The same pair assignments apply to natural questions and intact diagnostic quartets. Original weights average blocks equally, tasks equally within blocks, and option counts equally within tasks. We condition these weights on each group rather than recomputing a row-wise mean. Within a source split, scores average its three fixed LoRA fits; pooled scores weight the two splits equally. Both splits reuse the same questions and Base predictions. Pooled counts therefore denote source-reference memberships, not independent samples. Six omitted ineligible quartets remain omitted.

Natural gains increase across agreement groups in both splits (Table~\ref{tab:agreement-strata}). Pooled QAcc increases, although split-1 QAcc is not monotonic. Absolute agreement cutoffs have uneven task coverage, which motivates ranking target rules within each task/$K$ stratum before assigning agreement groups.

\begin{table*}[t]
\centering\small
\setlength{\tabcolsep}{4pt}
\begin{tabular*}{\textwidth}{@{\extracolsep{\fill}}llrrrrll@{}}
\toprule
\rowcolor{tableblue}\multicolumn{8}{c}{Natural questions}\\
Split & Agreement & Questions & Base & LoRA & Gain & 95\% gain CI & Conflict (\%)\\
\midrule
1 & Low & 1600 & 37.03 & 44.29 & +7.26 & [4.29, 10.31] & 52.50\\
1 & Middle & 1280 & 39.26 & 54.82 & +15.56 & [12.48, 18.55] & 41.67\\
1 & High & 1280 & 38.15 & 65.10 & +26.95 & [23.96, 29.86] & 34.83\\
\midrule
2 & Low & 1600 & 37.08 & 52.78 & +15.69 & [12.93, 18.33] & 46.82\\
2 & Middle & 1280 & 37.37 & 60.61 & +23.24 & [20.42, 26.24] & 39.84\\
2 & High & 1280 & 39.97 & 69.16 & +29.19 & [25.87, 32.47] & 28.06\\
\midrule
Pooled & Low & 3200 & 37.06 & 48.53 & +11.48 & [9.22, 13.59] & 49.66\\
Pooled & Middle & 2560 & 38.31 & 57.71 & +19.40 & [16.97, 21.89] & 40.76\\
Pooled & High & 2560 & 39.06 & 67.13 & +28.07 & [25.68, 30.69] & 31.45\\
\bottomrule
\end{tabular*}
\par\medskip
\begin{tabular*}{\textwidth}{@{\extracolsep{\fill}}llrrrrrl@{}}
\toprule
\rowcolor{tableblue}\multicolumn{8}{c}{Complete diagnostic quartets}\\
Split & Agreement & Quartets & Base & LoRA & Gain & QAcc & 95\% QAcc CI\\
\midrule
1 & Low & 100 & 36.66 & 39.26 & +2.60 & 5.58 & [2.75, 8.33]\\
1 & Middle & 78 & 35.48 & 43.49 & +8.01 & 3.17 & [1.05, 5.28]\\
1 & High & 76 & 31.53 & 52.27 & +20.74 & 10.98 & [7.20, 14.77]\\
\midrule
2 & Low & 98 & 32.68 & 44.06 & +11.38 & 6.89 & [4.84, 8.85]\\
2 & Middle & 78 & 36.75 & 55.08 & +18.33 & 15.67 & [12.00, 19.33]\\
2 & High & 78 & 35.61 & 59.53 & +23.92 & 17.15 & [12.23, 21.82]\\
\midrule
Pooled & Low & 198 & 34.67 & 41.66 & +6.99 & 6.24 & [4.52, 7.93]\\
Pooled & Middle & 156 & 36.11 & 49.28 & +13.17 & 9.42 & [7.22, 11.63]\\
Pooled & High & 154 & 33.57 & 55.90 & +22.33 & 14.07 & [11.33, 16.74]\\
\bottomrule
\end{tabular*}
\caption{Low-agreement pairs retain positive pooled transfer gains but lower joint correctness than high-agreement pairs. Accuracy and QAcc are percentages; gains and their intervals are percentage points. Diagnostic Base and LoRA columns give item accuracy, and QAcc gives LoRA all-four correctness. Base QAcc is zero in every group. Pooled counts retain both source-reference memberships for questions shared across the two source pools.}
\label{tab:agreement-strata}
\end{table*}

\lead{Absolute conflict scores distinguish transfer gains from their full-set contributions.}
Conflict gains differ in sign across source splits, whereas nonconflicts improve in both (Table~\ref{tab:conflict-absolute}). Table~\ref{tab:nearest-weight-details} retains their weighted full-set gain contributions.

\begin{table*}[t]
\centering\small
\setlength{\tabcolsep}{4pt}
\begin{tabular*}{\textwidth}{@{\extracolsep{\fill}}llrrrrrr@{}}
\toprule
\rowcolor{tableblue}
Split & Source--target answers & Questions & Weight (\%) & Base & LoRA & Readout & Gain (pp)\\
\midrule
1 & Conflict & 1907 & 43.73 & 34.49 & 32.07 & 34.02 & -2.43\\
1 & Agree & 2253 & 56.27 & 40.83 & 70.93 & 47.73 & +30.09\\
2 & Conflict & 1722 & 38.90 & 31.10 & 43.72 & 34.50 & +12.62\\
2 & Agree & 2438 & 61.10 & 42.49 & 70.74 & 48.83 & +28.25\\
\bottomrule
\end{tabular*}
\caption{Conflict-conditioned absolute accuracy exposes different transfer outcomes across the two source splits. Scores are percentages; gains compare LoRA with Base under the original weights conditioned on each subset. The readout column reports the MLP selected separately using the validation inputs of each source-rule split.}
\label{tab:conflict-absolute}
\end{table*}

\lead{Fully conflicting quartets have low pooled joint success.}
Conflict load counts target answers differing from native nearest-source references within an intact quartet. Fully conflicting quartets show opposite gain signs across splits and little pooled improvement (Table~\ref{tab:quartet-conflict-load}). Intermediate loads do not follow a monotonic gain trend in this benchmark.

\begin{table*}[t]
\centering\small
\setlength{\tabcolsep}{4pt}
\begin{tabular*}{\textwidth}{@{\extracolsep{\fill}}lrrrrrrrr@{}}
\toprule
\rowcolor{tableblue}
Split & Load & Quartets & Task/$K$ & Weight (\%) & Base & LoRA & Gain (pp) & QAcc\\
\midrule
1 & 0 & 37 & 7 & 13.14 & 32.93 & 47.29 & +14.36 & 4.88\\
1 & 1 & 19 & 4 & 11.22 & 46.43 & 58.33 & +11.90 & 14.29\\
1 & 2 & 129 & 10 & 50.32 & 39.17 & 48.89 & +9.71 & 7.15\\
1 & 3 & 30 & 4 & 11.75 & 24.09 & 41.74 & +17.65 & 4.55\\
1 & 4 & 39 & 8 & 13.57 & 20.47 & 14.37 & -6.10 & 0.00\\
\midrule
2 & 0 & 46 & 7 & 17.74 & 36.90 & 59.69 & +22.79 & 23.09\\
2 & 1 & 19 & 6 & 9.29 & 46.55 & 58.05 & +11.49 & 10.34\\
2 & 2 & 130 & 10 & 49.79 & 34.07 & 52.63 & +18.56 & 11.59\\
2 & 3 & 27 & 5 & 11.86 & 37.84 & 54.50 & +16.67 & 13.51\\
2 & 4 & 32 & 7 & 11.32 & 22.41 & 31.29 & +8.88 & 2.83\\
\midrule
Pooled & 0 & 83 & 7/7 & 15.44 & 34.91 & 53.49 & +18.58 & 13.99\\
Pooled & 1 & 38 & 4/6 & 10.26 & 46.49 & 58.19 & +11.70 & 12.32\\
Pooled & 2 & 259 & 10/10 & 50.05 & 36.62 & 50.76 & +14.14 & 9.37\\
Pooled & 3 & 57 & 4/5 & 11.81 & 30.96 & 48.12 & +17.16 & 9.03\\
Pooled & 4 & 71 & 8/7 & 12.45 & 21.44 & 22.83 & +1.39 & 1.42\\
\bottomrule
\end{tabular*}
\caption{All-four conflict yields little pooled accuracy gain and low joint correctness. Load counts conflicting members per complete quartet. Base and LoRA columns give diagnostic item accuracy; QAcc gives LoRA all-four correctness. Base QAcc is zero throughout. Weight is original diagnostic mass, and Task/$K$ counts represented strata, listed by source split in pooled rows. Pooled counts reuse the same quartets under two source references.}
\label{tab:quartet-conflict-load}
\end{table*}

Intervals use 2,000 paired resamples with seed 2026101201. Sampling units are shared natural states or complete diagnostic quartets within block and task/$K$. Joint cohort memberships under both source splits remain fixed, preserving shared observations across models and source references. Intervals condition on tested rules, fixed models, and observed cohorts. Exact-load cohorts include singleton resampling units, restricting uncertainty estimates. We report their finite-cohort scores and coverage as descriptive diagnostics of transfer under concentrated source--target answer conflict.

\subsection{Adjacent Randomized Jev Repetitions}
\label{app:adjacent-jev}
The schedule places six requests for each item adjacently: the original and two rewrites each occur twice in randomized order. Item blocks are also randomized. The 320 questions form 80 complete quartets balanced across the two existing splits. All 1,920 logical measurements completed with \texttt{jev-1.13.0}. Successful replacements count once per logical request. The repeated-call comparison uses the fixed development questions and the same original task labels and question identities across all six paired adjacent request conditions per item.

For item $i$, within-text disagreement $W_i$ averages the three identical-text pairs. Across-wording disagreement $B_i$ averages the twelve pairs combining distinct wordings. We first compute $B_i-W_i$ within item, average items within each split, and weight splits equally. The 960 within-text and 3,840 across-wording pairs share questions and are not independent samples. The 95\% interval uses 2,000 whole-quartet bootstrap draws stratified by split, with seed 2026100807 and all six responses retained together. Figure~\ref{fig:jev-repeat} reports within-text disagreement, across-wording disagreement, and their paired difference under shared quartet resampling.

\lead{The excess disagreement persists when request spacing and interruptions are examined.}
All five slot-lag strata have positive across-minus-within differences. Table~\ref{tab:jev-time} retains actual-time denominators, including sparse strata. A posthoc sensitivity analysis excludes an entire quartet if any item had a retry or any adjacent successful-request start gap within an item was at least 16 seconds. It removes five quartets and retains 75, with 39/36 in splits 1/2. Split weights remain equal. The retained-sample contrast and interval are shown beside the full result in Figure~\ref{fig:jev-repeat}. Whole quartets are excluded.

\begin{table*}[t]
\centering\small
\begin{tabular}{lcccc}
\toprule
\rowcolor{tableblue}
Actual start gap & Within pairs & Disagree (\%) & Across pairs & Disagree (\%)\\
\midrule
$<2$ s & 523 & 9.37 & 1991 & 35.26\\
2--4 s & 324 & 9.88 & 1363 & 35.95\\
4--8 s & 105 & 6.67 & 452 & 34.51\\
8--16 s & 0 & N/A & 2 & 100.00\\
$\ge16$ s & 8 & 12.50 & 32 & 37.50\\
\bottomrule
\end{tabular}
\caption{Cross-wording disagreement exceeds same-text disagreement in every elapsed-time stratum with both comparisons. The 8--16 s stratum has no same-text pairs; the counts identify the available paired coverage.}
\label{tab:jev-time}
\end{table*}

\subsection{Larger System and Boolean Policies}

\lead{A larger instruction-tuned system does not uniformly dominate the adapted model.}
Qwen3.5-9B remains below arithmetic-adapted LoRA when directly scoring candidate answers under plain-text and native-chat prompts. Native-template scoring disables thinking (Appendix~\ref{app:optional}). The larger system performs better on Boolean policies, with weak exception-handling QAcc (Appendix Table~\ref{tab:policy}). Finite-state transitions also yield limited natural accuracy and zero quartet success. These comparisons distinguish answer-interface sensitivity from transfer across different executable rule families.

\begin{table*}[t]
\centering\small
\setlength{\tabcolsep}{4pt}
\begin{tabular*}{\textwidth}{@{\extracolsep{\fill}}lcccccc@{}}
\toprule
\rowcolor{tableblue}
& \multicolumn{2}{c}{Priority policy} & \multicolumn{2}{c}{Exception policy} & \multicolumn{2}{c}{Finite-state transitions}\\
\cmidrule(lr){2-3}\cmidrule(lr){4-5}\cmidrule(lr){6-7}
System & Natural & QAcc & Natural & QAcc & Natural & QAcc\\
\midrule
Base & 60.00 & 0.00 & 31.87 & 0.00 & 25.31 & 0.00\\
Arithmetic LoRA & 51.88 & 0.00 & 43.12 & 2.50 & 21.88 & 0.00\\
Qwen3.5-9B & 78.12 & 22.50 & 76.25 & 2.50 & 32.81 & 0.00\\
\rowcolor{tablepeach} Always deny & 58.13 & 0.00 & 78.13 & 0.00 & \multicolumn{2}{c}{Not applicable}\\
\bottomrule
\end{tabular*}
\caption{The tested systems show different competence across Boolean policies and finite-state transitions. All systems have zero transition-task QAcc. Scores are descriptive finite-set percentages; the arithmetic LoRA is reused without new-task training and the always-deny baseline applies to both evaluated Boolean policy rule families.}
\label{tab:policy}
\end{table*}

The Qwen3.5-9B checkpoint revision is \texttt{c202236235762e1c871ad0}\allowbreak\texttt{ccb60c8ee5ba337b9a}. We score all 5,176 arithmetic questions in all three wordings, yielding 15,528 predictions, without training, chain-of-thought prompting, or checkpoint selection. The instruction-tuned system differs from the controlled backbone in more than parameter count. The interface uses the same prose template as the smaller backbone: State, Question, Options, and an Answer prefix, with no chat template or added special tokens. It left-pads inputs, applies the BF16 body, projects the final hidden state onto FP32 original candidate-letter rows, and selects the highest-scoring valid option. This configuration is direct scoring, rather than generated chain-of-thought or a chat interaction.

The Boolean extension uses six flags and twenty fixed policies: ten first-match priority policies and ten policies with one exception and a waiver. Each policy partitions the six flags into three two-literal conjunctions with randomized signs. Duplicate truth tables are rejected within family. Sixteen states per policy are sampled uniformly without replacement before labels are used. Priority semantics are first matching action among Allow, Deny, Allow, followed by default Deny. Exception semantics are $a\land(\neg b\lor c)$: the waiver restores the base rule and cannot grant permission alone. The fixed depth provides no depth-trend experiment. Exhaustive checking of all 64 flag states per policy gives 1,280 cross-oracle comparisons. Each family has 160 natural questions and 40 quartets. The natural and diagnostic sets do not overlap, but the diagnostic quartets share policy--state inputs (262 distinct diagnostic inputs among 320 rows). We therefore report finite-set descriptive results without row-IID significance tests. The arithmetic LoRA is the fixed split-1/state-101 checkpoint, with no policy training. Base, that LoRA, and 9B each produce 640 predictions, totaling 1,920. Always-deny accuracy makes the natural label imbalance explicit. A fixed answer cannot solve the checkerboard quartets.

\begin{takeawaybox}
Check each target rule family and answer interface when evaluating an adapted system. Report joint success and item accuracy together to distinguish reliable decisions from partial success.
\end{takeawaybox}

\subsection{Actual Wording and Policy Inputs}
The Boolean templates ask either for the first matching action in an ordered Allow/Deny list, or for a base permission with an exception and waiver. In the order credential, sponsor, training, hold, pass, audit, the first priority example has values $(0,0,0,1,0,1)$ and oracle Allow. The first exception example has $(1,0,0,0,0,0)$ and oracle Deny. The full semantics are specified above. Later examples give the serialized Jev request and the exact table changes applied to that same native request.

The arithmetic example is the first frozen item, selected independently of predictions. All three wordings share \texttt{State: \{"facts": \{"id":"R0","x":12,"y":8,"z":-12\}\}}, options \texttt{A) no} and \texttt{B) yes}, and the final \texttt{Answer:} prefix. The oracle is B. Only the question field changes below. Displayed line wrapping is typographic.

\begin{promptbox}{Original arithmetic question}
\footnotesize\ttfamily\raggedright Using facts, is (min(x, y) - (y + z)) strictly greater than 0? Select no or yes. Only x, y, z participate. Ignore other fields and notes.\par
\end{promptbox}

\begin{promptbox}{Equivalent functional question}
\footnotesize\ttfamily\raggedright Evaluate difference(minimum(x, y), sum(y, z)) on facts. Return yes exactly when its value exceeds 0; otherwise return no. Only x, y, z participate. Ignore other fields and notes. Here sum(a,b) means a+b, difference(a,b) means a-b, product(a,b) means a*b, minimum(a,b) means min(a,b), and maximum(a,b) means max(a,b).\par
\end{promptbox}

\begin{promptbox}{Equivalent verbal question}
\footnotesize\ttfamily\raggedright For the facts, decide whether (min(x, y) minus (y plus z)) is strictly positive. Choose yes for a result greater than 0, and no otherwise. Only x, y, z participate. Ignore other fields and notes.\par
\end{promptbox}

\subsection{Support of the Reassociation Reference}
\label{app:calibration-support}
The task-level reference uses separate cell/task/$K$/slot/gold strata. A singleton stratum cannot move. A stratum with constant predictions also cannot change that model's predictions, even when it contains several entries. The two measures overlap and must not be added. Table~\ref{tab:reassociation-support} reports counts and slot weights, with weights averaged equally over the six fixed cells. A slot weight is one quarter of its quartet's task-level macro weight. The reference uses 1,000 draws and seed 2026100802 for each cell, task, and model. Table~\ref{tab:selection-detail} exposes how the selection departure varies across cells and option counts. For reassociation draw $b$, define $d_b=100(Q_{\rm obs}-Q_{{\rm ref},b})$. The table gives $\bar d$ and the 2.5th and 97.5th percentiles of $d_b$ in percentage points. The bracket is a descriptive difference range, not a range of raw QAcc values or a confidence interval. These finite-cohort differences are unadjusted for multiplicity and conditional on the evaluated rule and state pairs.
{\small
\begin{table*}[t]
\centering
\begin{tabular}{llcccccc}
\toprule
\rowcolor{tableblue}
Task & Model & Strata & Size range & Singletons & Weight (\%) & Constant & Weight (\%)\\
\midrule
Binary & Base & 24 & 8--8 & 0 & 0.00 & 0 & 0.00\\
Binary & Jev & 24 & 8--8 & 0 & 0.00 & 12 & 50.00\\
Binary & LoRA & 24 & 8--8 & 0 & 0.00 & 5 & 20.83\\
Routing & Base & 262 & 1--6 & 102 & 17.71 & 157 & 47.05\\
Routing & Jev & 262 & 1--6 & 102 & 17.71 & 130 & 30.73\\
Routing & LoRA & 262 & 1--6 & 102 & 17.71 & 129 & 29.34\\
Rating & Base & 172 & 1--8 & 40 & 6.94 & 165 & 93.40\\
Rating & Jev & 172 & 1--8 & 40 & 6.94 & 71 & 25.17\\
Rating & LoRA & 172 & 1--8 & 40 & 6.94 & 69 & 27.08\\
Selection & Base & 240 & 1--6 & 80 & 13.89 & 159 & 57.12\\
Selection & Jev & 240 & 1--6 & 80 & 13.89 & 112 & 29.69\\
Selection & LoRA & 240 & 1--6 & 80 & 13.89 & 108 & 26.74\\
\bottomrule
\end{tabular}
\caption{Small and constant-prediction strata constrain the variation available to reassociation. Counts pool disjoint strata across six cells; weights average cell-level task macro weights. Constant counts include singletons.}\label{tab:reassociation-support}
\end{table*}}
{\small
\begin{table*}[t]
\centering
\begin{tabular}{llcccc}
\toprule
\rowcolor{tableblue}
Split/state & $K$ & LoRA QAcc & LoRA $\bar d$ [$q_{.025},q_{.975}$] & Jev QAcc & Jev $\bar d$ [$q_{.025},q_{.975}$]\\
\midrule
1/101 & 2 & 50.00 & +21.48 [+0.00, +37.50] & 25.00 & +2.75 [-12.50, +25.00]\\
1/101 & 4 & 0.00 & +0.00 [+0.00, +0.00] & 0.00 & +0.00 [+0.00, +0.00]\\
1/101 & 8 & 0.00 & +0.00 [+0.00, +0.00] & 12.50 & +9.09 [+0.00, +12.50]\\
1/202 & 2 & 25.00 & +11.99 [-12.50, +25.00] & 12.50 & +8.77 [+0.00, +12.50]\\
1/202 & 4 & 0.00 & -2.69 [-12.50, +0.00] & 12.50 & +10.86 [+0.00, +12.50]\\
1/202 & 8 & 0.00 & +0.00 [+0.00, +0.00] & 12.50 & +10.90 [+0.00, +12.50]\\
1/303 & 2 & 25.00 & +11.21 [+0.00, +25.00] & 12.50 & -2.54 [-25.00, +12.50]\\
1/303 & 4 & 25.00 & +9.15 [-12.50, +25.00] & 12.50 & +11.66 [+0.00, +12.50]\\
1/303 & 8 & 0.00 & -2.76 [-12.50, +0.00] & 0.00 & -0.78 [-12.50, +0.00]\\
2/101 & 2 & 12.50 & +3.33 [-12.50, +12.50] & 25.00 & +14.96 [+0.00, +25.00]\\
2/101 & 4 & 0.00 & -3.09 [-12.50, +0.00] & 25.00 & +6.54 [+0.00, +12.50]\\
2/101 & 8 & 12.50 & +4.28 [+0.00, +12.50] & 0.00 & +0.00 [+0.00, +0.00]\\
2/202 & 2 & 25.00 & +2.73 [-12.50, +12.50] & 50.00 & -3.04 [-12.50, +0.00]\\
2/202 & 4 & 12.50 & +10.88 [+0.00, +12.50] & 12.50 & +3.86 [-12.50, +12.50]\\
2/202 & 8 & 0.00 & -0.16 [+0.00, +0.00] & 0.00 & -0.83 [-12.50, +0.00]\\
2/303 & 2 & 12.50 & +5.09 [-12.50, +12.50] & 12.50 & +5.60 [-12.50, +12.50]\\
2/303 & 4 & 25.00 & +20.89 [+0.00, +25.00] & 0.00 & -7.09 [-25.00, +0.00]\\
2/303 & 8 & 12.50 & +8.43 [+0.00, +12.50] & 12.50 & +9.34 [+0.00, +12.50]\\
\bottomrule
\end{tabular}
\caption{Maximum-selection departures vary across cells and option counts. QAcc is in percentages and mean differences $\bar d$ in pp; brackets give quantiles of $d_b$, with eight complete diagnostic quartets in every cell/$K$ stratum.}\label{tab:selection-detail}
\end{table*}}

\subsection{Answer-Transition Weighting Sensitivity}
\label{app:weighting}
Let $N$ count applicable conflicts between the original oracle and altered-table target. Let $B$ count conflicts answered correctly with the correct table, and $T$ count paired transitions to the altered-table target. We report both $T/N$ and $T/B$. Micro weighting pools counts. Task/$K$ weighting pools cells within a task/$K$, averages equally over available $K$ within each task, then over tasks. Equal-cell weighting first computes the same task/$K$ macro within each cell, then averages cells. Zero-denominator strata are excluded from the relevant rate and their coverage is recorded. All applicable Jev strata have defined conditional rates, with 20 cell/task/$K$ strata for row and target interventions and 12 for column interventions. Single-record tasks have no nonidentity column permutation and are inapplicable.

{\small
\begin{table*}[t]
\centering
\begin{tabular}{lcccccc}
\toprule
\rowcolor{tableblue}
Intervention & \multicolumn{2}{c}{Micro} & \multicolumn{2}{c}{Task/$K$ macro} & \multicolumn{2}{c}{Equal cell}\\
 & $T/N$ & $T/B$ & $T/N$ & $T/B$ & $T/N$ & $T/B$\\
\midrule
Formula rows & 76.82 & 88.48 & 78.81 & 86.47 & 79.18 & 86.58\\
Record columns & 91.05 & 97.74 & 91.10 & 97.46 & 91.08 & 97.41\\
Target values & 75.94 & 86.17 & 77.34 & 84.62 & 77.34 & 84.35\\
\bottomrule
\end{tabular}
\caption{Jev\textquotesingle{}s directed transitions persist under micro, task-balanced, and equal-cell weighting. All weighting schemes reuse the same paired requests and archived integer transition counts, changing their aggregation rather than the observed answers or the conflict questions entering the archived counts of paired answer transitions.}\label{tab:transition-weights}
\end{table*}}

\begin{figure*}[t]
\centering
\includegraphics[width=\linewidth]{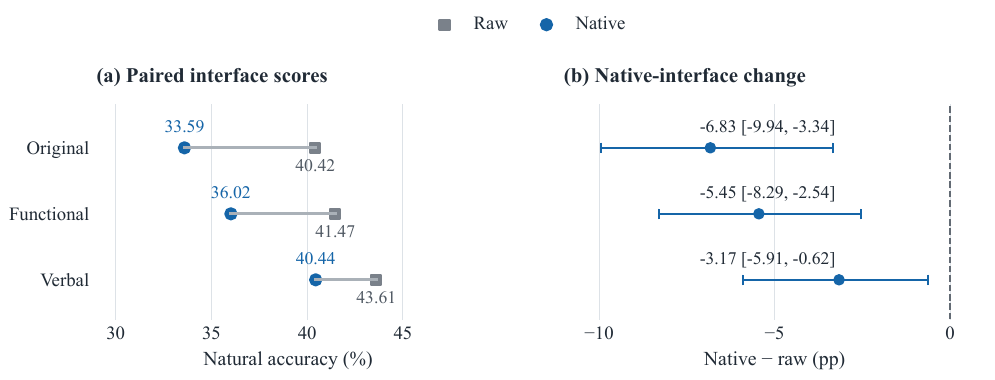}
\caption{The fixed native chat template lowers 9B natural accuracy under all three wordings. (a) Lines connect raw and native scores on paired inputs. (b) Paired native--raw differences are negative, with 95\% intervals from resampling the 13 reserved blocks. The score axis in (a) is restricted to 29--49\% for resolving paired changes.}
\label{fig:native-interface}
\end{figure*}

\subsection{Native Jev Requests and Table Interventions}
\label{app:jev-native}
Figures~\ref{fig:request-1} and~\ref{fig:request-2} show the original instruction and complete-value native request. The original shares the model, criteria, type, and records shown in the complete-value request, but omits supplementary scores. Both full payloads are retained in \path{examples/native_requests.json}. Formatting preserves every string value. They were sent to \url{https://api.typesafe.ai/v1/systemone}. The \texttt{state} field is itself a serialized JSON string. The decision is requested through \texttt{questions.decision}, not through the local models' letter-answer prompt. Successful responses identify \texttt{jev-1.13.0}. The example is the first two-record routing item in split 1/state 101, chosen independently of predictions. The example specifies the evaluated interface and exact interventions; aggregate transition rates are reported over the full applicable conflict set for each table intervention.

\noindent\textbf{D0: original request.}

\begin{figure}[t]
\centering
\begin{promptbox}{Original decision: instruction}
\begin{Verbatim}[fontsize=\footnotesize,breaklines=true,breakanywhere=true]
Select the record with the smallest routing penalty, computed as ((x * y) + max(y, z)). If several records tie, select the one with the smallest numeric record ID. Only x, y, z participate. Ignore other fields and notes.
\end{Verbatim}
\end{promptbox}
\caption{The original native instruction asks for the minimum-scoring record and specifies the tie-break. The evaluated native request string is retained exactly.}
\label{fig:request-1}
\end{figure}

\noindent\textbf{D2: correct all-rule table.}
D2 adds all formula values and replaces the D0 instruction to ignore supplementary fields with an instruction explaining how to use them. D2 and D3 share this instruction. All subsequent mapping and target interventions preserve the D2 question exactly. Punctuation inside these archived strings is retained verbatim.

\begin{figure*}[t]
\centering
\begin{promptbox}{Complete-value decision: native service request}
\begin{Verbatim}[fontsize=\footnotesize,breaklines=true,breakanywhere=true]
{
  "model": "jev-latest",
  "questions": {"decision": {"criteria": {"R0": null, "R1": null}, "type": "choice",
    "instructions": "Select the record with the smallest routing penalty, computed as ((x * y) + max(y, z)). If several records tie, select the one with the smallest numeric record ID. Only x, y, z participate. Supplementary information, when present, provides computed scores for formulas or a semantic answer. Use values for the exact formula in this question, then apply its comparison, thresholds, and tie-break. A supplied semantic answer must be matched to the displayed options. An unavailable entry supplies no answer or value; use the original facts or records in that case."}},
  "state": "{\"records\": [{\"id\": \"R0\", \"x\": 3, \"y\": 2, \"z\": -8}, {\"id\": \"R1\", \"x\": 4, \"y\": -2, \"z\": 7}], \"supplementary_computed_scores\": [{\"formula\": \"((x * y) + max(y, z))\", \"record_scores\": {\"R0\": 8, \"R1\": -1}}, {\"formula\": \"((x * y) - (y - z))\", \"record_scores\": {\"R0\": -4, \"R1\": 1}}, {\"formula\": \"((x + y) * (y - z))\", \"record_scores\": {\"R0\": 50, \"R1\": -18}}, {\"formula\": \"((x + y) * max(y, z))\", \"record_scores\": {\"R0\": 10, \"R1\": 14}}, {\"formula\": \"((x + y) - (y * z))\", \"record_scores\": {\"R0\": 21, \"R1\": 16}}, {\"formula\": \"max((x * y), (y * z))\", \"record_scores\": {\"R0\": 6, \"R1\": -8}}, {\"formula\": \"max((x - y), (y * z))\", \"record_scores\": {\"R0\": 1, \"R1\": 6}}, {\"formula\": \"min((x - y), (y * z))\", \"record_scores\": {\"R0\": -16, \"R1\": -14}}]}"
}
\end{Verbatim}
\end{promptbox}
\caption{The complete-value request supplies scores for every rule while retaining the decision task. The displayed strings preserve the evaluated request, including its instruction, records, formula labels, and numeric values.}
\label{fig:request-2}
\end{figure*}

\noindent\textbf{Worked interventions on the same request.}
The requested formula is $(x y)+\max(y,z)$. Its correct values are $(8,-1)$ for $(\mathrm{R0},\mathrm{R1})$, so the original oracle selects R1. The row intervention cyclically moves every score row forward by two positions while retaining formula labels. The requested row becomes $(50,-18)$ and still selects R1, so this item is \emph{not} a row-conflict case. The column intervention swaps R0 and R1 values in every row, giving $(-1,8)$ in the requested row and a new target R0. The target intervention changes only the requested row's R0 value from $8$ to $-1$. The values then tie at $(-1,-1)$ and the fixed record-ID tie-break selects R0. In both conflict cases the original task's gold answer remains R1. This example separates changes in the table from changes in its implied answer and identifies the conflicts that enter every denominator in this paired-transition analysis.

\begin{table*}[t]
\centering\small
\setlength{\tabcolsep}{5pt}
\begin{tabularx}{\textwidth}{@{}lLL@{}}
\toprule
\rowcolor{tableblue}
Evaluation & Controlled factor & Evaluation setting\\
\midrule
CheckList / contrast sets & Local input changes & Behavioral and contrastive tests\\
LogicBench & Logical inference patterns & Three kinds of logic\\
RuleArena & Rules and tool assistance & Real-world regulations\\
CFQ / COGS / Ordered CommonGen & Combinations and concept order & Compositional generalization\\
Mind Your Format / symbol binding & Templates and option symbols & Answer-interface variation\\
Cacheable decision models & Joint versus independent encoding & Rule-following decision scores\\
\midrule
\rowcolor{tablepeach} This work & Rules, states, wording, value mappings & Oracle-checked decision diagnostics\\
\bottomrule
\end{tabularx}
\caption{Controlled input changes provide different kinds of evidence across evaluation designs. Our oracle-checked setting crosses rules and states and intervenes on equivalent wording and supplied-value mappings.}
\label{tab:literature-comparison}
\end{table*}

\subsection{Additional Rule Split and Interface Extensions}
\label{app:optional}
For finite-state symbol-mapping swaps, correct-to-new-target transitions are 9/88 (10.23\%) for Base, 10/82 (12.20\%) for LoRA, and 23/140 (16.43\%) for Qwen3.5-9B. The denominator contains initially correct conflicting questions. Table~\ref{tab:policy} reports the natural and quartet scores.
\lead{A further disjoint rule split reproduces an adaptation gain dominated by nonconflict questions.}
The added split uses 16 components disjoint from the original source/development pools, state seed 101, 5,760 training rows, and 1,920 evaluation rows. The original LoRA recipe runs 720 updates, with source-validation NLL selecting epoch 2. These components had appeared in reserved evaluation, so the result is development replication: three rule splits and seven cells overall. Natural accuracy / QAcc are 38.80\% / 0.00\% for Base and 58.46\% / 19.79\% for LoRA. The decomposition uses full-set weights. Conflict weight is 36.98\%, with conditional gains of 5.99 pp versus 27.69 pp on agreeing answers.

\lead{The fixed native template does not improve the larger model's direct decision scores.}
The native check places each rendered prompt in one user message through Qwen3.5-9B's chat template with \texttt{enable\_thinking=False}. It directly scores candidate letters for all 5,176 questions in each wording. Figure~\ref{fig:native-interface} compares this with raw prompting. Paired intervals use 2,000 resamples of the 13 existing blocks. The comparison holds model weights and question identities fixed, using direct candidate scoring without generated reasoning or additional selection of the input prompt or model checkpoint.

\lead{The tested arithmetic adaptation does not confer broad finite-state rule competence.}
The sequential extension uses four states, two event symbols, and 20 noncommuting transition tables. Each event permutes the states. We enumerate ordered pairs of distinct permutations whose composition differs in at least one state when reversed, shuffle with seed 2026100807, and retain 20 tables. Every start state and X/Y sequence of length two to four gives 112 inputs per table, with length probabilities $1/7,2/7,4/7$ under uniform sampling. Each natural set draws 16 inputs without replacement.

Adjacent tables form ten pairs. Diagnostic construction excludes both tables' natural inputs, groups remaining inputs by their two distinct answers, and draws eight quartets per pair from opposite-answer groups. Inputs may recur across quartets within a pair. Each quartet receives a single shared permutation of the displayed options.

\noindent Independent function composition verifies every table/input oracle. The resulting 320 natural questions and 80 quartets total 640 questions.
\noindent Correct and X/Y-swapped tables are evaluated with Base, split-1/state-101 arithmetic LoRA, and 9B, producing 3,840 predictions without new-task training. The 449 conflicts pool natural and diagnostic inputs. Transition rates condition on initially correct conflicts, so denominators differ by model. All systems have zero QAcc, which alone does not establish an additional structural deficit. Intervals resample the ten table pairs.

\section{Comparison of Evaluation Designs}
\label{app:related-designs}
Table~\ref{tab:literature-comparison} compares which input factors each evaluation controls and the rule-following settings in which those behavioral input controls are applied. Local perturbation tests reveal behavioral sensitivity, while compositional benchmarks test generalization across combinations. Our quartets make rule and state identities explicit, so item accuracy and all-four correctness answer different evaluation questions. Value-mapping interventions further distinguish following supplied values from retaining the original answer. These designs can complement one another when their controlled factors and score denominators are transparently reported together.

\end{document}